\documentclass[nolayout]{article}
\usepackage[utf8]{inputenc}
\usepackage[english]{babel}
\usepackage{authblk}
\usepackage{subcaption}
\usepackage{color, colortbl, xcolor}
\usepackage{dsfont}
\usepackage{amssymb,amsmath,amsthm}
\usepackage[pdftex]{graphicx}
\usepackage{footnote}
\usepackage{url}
\usepackage{times,dsfont}
\usepackage{fancyhdr}
\usepackage{geometry}
\usepackage[shortlabels]{enumitem}
\usepackage{breakcites}
\newcommand{\Nbb}{{\mathbb{N}}}
\newcommand{\Zbb}{{\mathbb{Z}}}
\newcommand{\window}{w}
\usepackage{multirow}
\usepackage{natbib}

\usepackage{hyperref}
\usepackage{url}
\usepackage{graphicx}

\title{Variational quantization for state space models}
\date{}

\author[1,2]{\'Etienne David}
\author[2]{Jean Bellot}
\author[3]{Sylvain Le Corff}

\affil[1]{Samovar, Département CITI, Telecom SudParis, Evry, 91011, France}
\affil[2]{Heuritech, Paris, 75003, France}
\affil[3]{LPSM, Sorbonne Universit\'e, UMR CNRS 8001, 75005, Paris}

\begin{document}

\maketitle

\begin{abstract}
Forecasting tasks using large datasets gathering thousands of heterogeneous time series is a crucial statistical problem in numerous sectors. The main challenge is to model a rich variety of time series, leverage any available external signals and provide sharp predictions with statistical guarantees. In this work, we propose a new forecasting model that combines discrete state space hidden Markov models with recent neural network architectures and training procedures inspired by vector quantized variational autoencoders. We introduce a variational discrete posterior distribution of the latent states given the observations and a two-stage training procedure to alternatively train the parameters of the latent states and of the emission distributions. By learning a collection of emission laws and temporarily activating them depending on the hidden process dynamics, the proposed method allows to explore large datasets and leverage available external signals. We assess the performance of the proposed method using several datasets and show that it outperforms other state-of-the-art solutions.
\end{abstract}

%{\bf Keywords:} Hidden Markov Model, Identifiability, Consistency, Expectation-Maximization, Fashion time series

\section{Introduction}

An increasingly common time series forecasting problem concerns the forecast of large datasets gathering thousands of heterogeneous sequences, see \cite{makridakis2018,makridakis2022,lai2018,zhou2021,david2022} and the references therein. One of the main difficulties is to design mathematical models for a large variety seasonal patterns, noise levels, trends and non-stationary changes. Additionally, some time-series datasets provide external signals that can be exploited to detect behaviors in the main time series that would otherwise be missed \citep{david2022, david2023}. Regarding this new type of forecasting use case, state-of-the-art solutions do not provide satisfactory results yet.

Parametric statistical models  have been largely studied during the past decades, see for instance \cite{box2015,hyndman2018}. Based on a sharp modeling of the time series distribution, these models can compute accurate predictions along with confidence intervals that make them largely used in numerous applications. Depending on the nature of the use case, many approaches have been proposed. The exponential smoothing model \citep{brown1961}, the Trigonometric Box-Cox transform, ARMA errors, Trend, and Seasonal components model (TBATS) \citep{alysha2011}, or the ARIMA model with the Box-Jenkins approach \citep{box2015} are for instance very popular parametric generative models. However, they cannot be used for large datasets gathering thousands of time series. As a new model need to be trained for each new time series, the training process can take considerable time depending on the number of sequences. Furthermore, much of the parametric models proposed cannot include external signals in their framework as the exact dependencies between the additional signals and the main ones remain unknown.

Hidden Markov models are other widespread models that have been largely studied in the literature \citep{sarkka2013,douc2014,chopin2020}. Introduced in the late 1960s, these generative models rely on hidden processes to describe the distribution of the target time series. Numerous variations have been proposed to fit different use cases \citep{juang1985,douc2004,touron2019}. In addition to providing accurate predictions, these models are supported by solid theoretical results on their identifiability and their consistency, see for instance \cite{douc2011,gassiat2016,gassiat2019} and references therein. However, when large datasets are considered, as a hidden state model has to be trained on each new time series, they are not well suited to forecast large samples gathering thousands of time series. Nevertheless, several contributions introducing hidden Markov models able to leverage external signals have been proposed, see \cite{bengio1994,radenen2012,Gonzalez2005,david2023}.

Finally, with recent improvements in speech processing and image recognition, neural-network-based models have emerged as the new state-of-the-art in time series forecasting. Among them, recurrent neural networks or sequence to sequence deep learning architectures \citep{hochreiter1997,vaswani2017} offer very appealing alternatives to exploit large time series dataset and leverage any kind of external signals. The DeepAR methods \citep{salinas2020}, N-HiTS and N-BEATS frameworks \citep{oreshkin2019,challu2023} and the following Transformer-based approaches \cite{lim2021,zhou2021,zhou2022,woo2022,wu2022,liu2022,woo2023,wu2023,nie2023} are examples of neural-network-based models that have obtained unprecedented accuracy levels in various applications. However, predictions computed by these methods are not interpretable and only a handful of theoretical results have been provided with these architectures.

In this paper, we introduce a new forecasting method combining hidden Markov models with recent neural-networks-based models. In this framework, it is assumed that time series are ruled by hidden Markov processes modeling the internal state of the time series. Depending on the hidden states dynamics, several emission laws are learned and specialized at forecasting specific types of behaviours. Maximum likelihood approaches cannot be used directly to train such a model and Expectation-Maximization (EM) algorithm is commonly used in this case. However, this algorithm is computationally costly, requires a fair amount of tuning and is very sensitive to the initialization. Thus, inspired by ideas brought with the Vector quantized VAE model \citep{oord2018}, a training process based on the Evidence Lower BOund (ELBO) learning alternatively the parameters of the latent states and of the emission distributions is introduced. On a collection of reference datasets, our approach outperforms current state-of-the-art solutions. We show that the model can forecast non-stationary time series, in particular when relevant external signals are included in the hidden states and the emission laws.

The paper is organized as follows. The proposed model is presented in Section~\ref{sec:nextmodel} along with the training procedure. Then, a complete experimental study is provided in Section~\ref{sec:nextresult} where the proposed framework is applied on several datasets, its accuracy assessed and evaluated in comparison with a collection of other state-of-the-art methods. Finally,  some research perspectives are given in Section~\ref{sec:nextconclusion}.

\section{Model and training procedure}
\label{sec:nextmodel}

\subsection{Model formulation}

\begin{figure*}
\centering
\includegraphics[width=1.\linewidth]{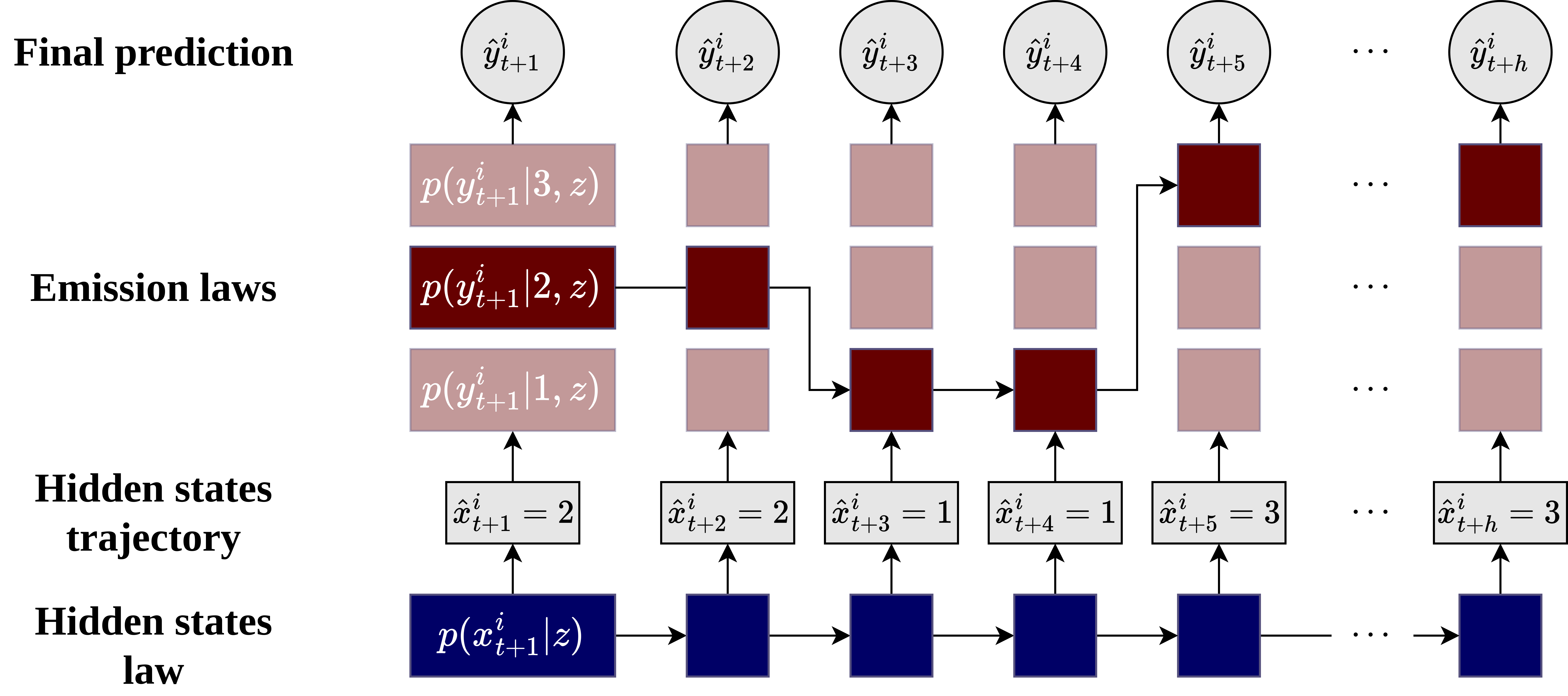}
\caption{Illustration of the proposed framework with 3 hidden states. Given the past of a time series $y^i_{t-\window+1:t}$ and possible additional external signals $w^i_{t-\window+1:t}$ (called $z$ in the figure), a trajectory of the hidden state $\hat{x}^i_{t+1:t+h}$ is drawn using the law of the hidden states. Then, conditioned on the values taken by the hidden process, one of the emission law is activated and used to compute the final prediction $\hat{y}^i_{t+1:t+h}$.}
\label{fig:nextframework}
\end{figure*}

Consider a dataset gathering $N\in\Nbb^*$ time series. For $i \in \{1,\cdots,N\}$, let $(y^i_t)_{t\in\Zbb}$ be the observation of the sequence $i$ and $(w^i_t)_{t\in\Zbb}$ a sequence of additional signals. These auxiliary variables may account for the history of some additional time series, or any other available information. The aim of the proposed model is to forecast, for all $i \in \{1,\cdots,N\}$, the next $h\geq1$ values of $y^i$ based on the past $\window\geq1$ values of $y^i$ and $w^i$, i.e. to estimate $p_{\theta}(y^i_{t+1:t+h}|y^i_{t-\window+1:t},w^i_{t-\window+1:t})$ the probability density function of the time series when the parameter value is $\theta$. We assume the existence of an additional discrete hidden process denoted by $(x^i_t)_{T+1\leq t \leq T+h}$ taking value in $\mathsf{X} = \{1,\cdots,K\}$ and that rules the density of $y^i_{t+1:t+h}$. This discrete hidden signal can be interpreted as a state or a regime in which is a sequence $i$ is at a time t. Depending on the values taken, $K$ different predictions can be computed for a same time series, all representing behaviours linked to the hidden regime of the time series. Thus, the previous density can be written as follows:
\begin{align*}
    p_{\theta}&(y^i_{t+1:t+h}|y^i_{t-\window+1:t},w^i_{t-\window+1:t}) \\
    &\hspace{1cm}=\sum_{x^i_{t+1:t+h}\in \mathsf{X}^p} p_{\theta}(y^i_{t+1:t+h},x^i_{t+1:t+h}|y^i_{t-\window+1:t},w^i_{t-\window+1:t}) \\
    &\hspace{1cm}= \sum_{x^i_{t+1:t+h}\in \mathsf{X}^p} \prod_{s=1}^h p_{\theta_{y}}(y^i_{t+s}|x^i_{t+1:t+s},y^i_{t-\window+1:t+s-1},w^i_{t-\window+1:t}) \\
    &\hspace{5cm}\times p_{\theta_{x}}(x^i_{t+s}|x^i_{t+1:t+s-1},y^i_{t-\window+1:t+s-1},w^i_{t-\window+1:t})\;,
\end{align*}
with the convention $p_{\theta}(.|x^i_{t+1:t+s-1},y^i_{t-\window+1:t+s-1},w^i_{t-\window+1:t}) = p_{\theta}(.|y^i_{t-\window+1:t},w^i_{t-\window+1:t})$ for $s=1$. Note that we decomposed the unknown parameter $\theta = (\theta_x,\theta_y)$ with i) the parameters corresponding to the law of the hidden states denoted by $\theta_x$ and ii) the parameters corresponding to the law of the main signal conditionally to the hidden states denoted by $\theta_y$. We consider the following assumptions. 
\begin{itemize}
    \item For all $i\in\{1,\cdots,H\}$ and all $s\in\{1,\cdots,h\}$, we assume that the conditional  law of $y^i_{t+s}$ depends on the current value of the external signal $x^i_{t+s}$ and the window $(y^i_{t-\window+1:t},w^i_{t-\window+1:t})$. 
    \item For all $i\in\{1,\cdots,H\}$ and all $s\in\{1,\cdots,h\}$, we assume that the conditional law of $x^i_{t+s}$ dependss on the previous $x^i_{t+s-1}$ and the window $(y^i_{t-\window+1:t},w^i_{t-\window+1:t})$. 
\end{itemize}
Thus, the predictive distribution can be written as follows.
\begin{multline}
\label{eq:modeldefinition}
    p_{\theta}(y^i_{t+1:t+h}|y^i_{t-\window+1:t},w^i_{t-\window+1:t}) \\= \sum_{x^i_{t+1:t+h}\in \mathsf{X}^p} \prod_{s=1}^h p_{\theta_{y}}(y^i_{t+s}|x^i_{t+s},y^i_{t-\window+1:t},w^i_{t-\window+1:t})p_{\theta_{x}}(x^i_{t+s}|x^i_{t+s-1},y^i_{t-\window+1:t},w^i_{t-\window+1:t})\;.
\end{multline}
The proposed framework is therefore a generative model composed of two parts: the law of the hidden process and the conditional emission laws of the main signal. An illustration of the proposed model is presented in Figure~\ref{fig:nextframework}.

\subsection{Training}
As $(x^i_t)_{t\in\Zbb}$ is never observed, the log likelihood is no longer computable for the proposed model and specific losses and algorithms have to be used. A relevant approach is to substitute the log likelihood by the Evidence Lower BOund (ELBO). Below, we recall the ELBO formulation adapted to the forecasting task studied in this work. For greater clarity, we omit the dependencies on $y^i_{t-\window+1:t},w^i_{t-\window+1:t}$. For all $i\in\{1,\cdots,N\}$,
$$
\log p_{\theta}(y^i_{t+1:t+h}) \geq \mathbb{E}_{x\sim q_{\phi}}\left[\log\frac{p_{\theta}(y^i_{t+1:t+h},x^i_{t+1:t+h})}{q_{\phi}(x^i_{t+1:t+h}|y^i_{t+1:t+h})} \right]\,,
$$
where the right hand side term defines the ELBO and $q_{\phi}(x^i_{t+1:t+h}|y^i_{t+1:t+h})$ the posterior variational probability of the hidden sequence $x^i_{t+1:t+h}$ conditioned on the observed sequence $y^i_{t+1:t+h}$.Therefore, the proposed loss function $\mathcal{L}(\phi,\theta_x,\theta_y)$ used to train the model is given by
\begin{align*}
\mathcal{L}(\phi,\theta_x,\theta_y)&= \frac{1}{N}\sum_{i=1}^N L^i(\phi,\theta_x,\theta_y)\\
&= \frac{1}{N}\sum_{i=1}^N \mathbb{E}_{x\sim q_{\phi}}\left[\log \frac{p_{\theta}(y^i_{t+1:t+h},x^i_{t+1:t+h}|y^i_{t-\window+1:t},w^i_{t-\window+1:t})}{q_{\phi}(x^i_{t+1:t+h}|y^i_{t-\window+1:t+h},w^i_{t-\window+1:t})} \right] \\
&= \frac{1}{N}\sum_{i=1}^N \mathcal{L}^i(\phi,\theta_y) + \mathcal{L}^i(\phi,\theta_x) - \mathcal{L}^i(\phi)\;,
\end{align*}
with 
\begin{align*}
    \mathcal{L}^i(\phi,\theta_y) &= \mathbb{E}_{x\sim q_{\phi}}\left[\sum_{s=1}^{h}\log p_{\theta_y}(y^i_{t+s}|x^i_{t+s},y^i_{t-\window+1:t},w^i_{t-\window+1:t})\right]\\
    \mathcal{L}^i(\phi,\theta_x) &= \mathbb{E}_{x\sim q_{\phi}}\left[\sum_{s=1}^{h}\log p_{\theta_x}(x^i_{t+s}|x^i_{t+s-1},y^i_{t-\window+1:t},w^i_{t-\window+1:t})\right]\\
    \mathcal{L}^i(\phi) &= \mathbb{E}_{x\sim q_{\phi}}\left[\sum_{s=1}^{h}\log q_{\phi}(x^i_{t+s}|y^i_{t-\window+1:t+h},w^i_{t-\window+1:t}) \right]\;.
\end{align*}

Inspired by ideas brought with Vector quantized VAE model \citep{oord2018}, the ELBO loss presented above is optimized in two steps. Whilst training the emission laws and the posterior variational law, the prior is not learnt and kept constant and uniform. After the convergence of the emission laws, they are frozen and the prior model is trained, guided by the learned posterior variational law.

\subsection{Implementation}

For all $i\in \{1,\cdots,N\}$ and $s\in \{1,\cdots,h\}$ consider the following assumptions.
\begin{itemize}
    \item Inspired by the DeepAR method introduced in \cite{salinas2020}, we assume that the $K$ emission laws $(p_{\theta_{y}}(y^i_{t+s}|x^i_{t+s}=k,y^i_{t-\window+1:t},w^i_{t-\window+1:t}))_{1\leq k\leq K}$ are Gaussian laws parameterized by $K$ different neural networks components. In fact, for each hidden state $k \in \{1,\cdots,K\}$, a neural-network-based model denoted $f_{\theta_y}^k$ is trained to predict $h$ couples of parameters for the Gaussian emission laws linked to the hidden regime $k$: $((\mu^{k,i}_{t+s},\sigma^{k,i}_{t+s}))_{1\leq s\leq h}$ with $\mu$ the mean and $\sigma$ the standard deviation.
	\begin{align*}
		p_{\theta_{y}}(y^i_{t+s}|x^i_{t+s}=k,y^i_{t-\window+1:t},w^i_{t-\window+1:t}) &= \mathcal{N}(y^i_{t+s};\mu^{k,i}_{t+s},\sigma^{k,i}_{t+s}) \\
		(\hat{\mu}^{k,i}_{t+s},\hat{\sigma}^{k,i}_{t+s}) &= f^k_{\theta_y}(y^i_{t-\window+1:t},w^i_{t-\window+1:t})_{s}
	\end{align*}
Note that the output of $f_{\theta_y}^k$ is a vector of Gaussian parameters. Thus, for all $s,s' \in \{1,\cdots,h\}$, the Gaussian parameters used to sample $p_{\theta_{y}}(y^i_{t+s}|x^i_{t+s}=k,$ $y^i_{t-\window+1:t},w^i_{t-\window+1:t})$ and $p_{\theta_{y}}(y^i_{t+s'}|x^i_{t+s'}=k,y^i_{t-\window+1:t},w^i_{t-\window+1:t})$ are calculated by the same neural-network-based model.
    \item The prior law of the hidden states is provided by a neural network component called $f_{\theta_x}$. Based on $y^i_{t-\window+:t+s-1},w^i_{-t-\window+1:t}$, $f_{\theta_x}$ returns the initial law and transition matrices of the hidden process: 
    \begin{align*}
        &p_{\theta_{x}}(x^i_{t+s}=k|x^i_{t+s-1}=j,y^i_{t-\window+:t},w^i_{-t-\window+1:t}) = \alpha^{i,j,k}_s\,,\\
        &\hat{\alpha}^{i,j,k}_s = f_{\theta_x}(y^i_{t-\window+1:t},w^i_{t-\window+1:t})_{s,j,k}\,.
    \end{align*}
    \item Finally, the posterior variational law of the hidden states $q_{\phi}(x^i_{t+s}|y^i_{t-\window+1:t+h},w^i_{t-\window+1:t})$ is learnt by a neural-network-based model named $f_{\phi}$. Based on $y^i_{t-\window+:t+s-1}$ and $w^i_{-t-\window+1:t}$, $f_{\phi}$ returns a matrix with $K\times h$ hidden state probabilities :
    \begin{align*}
        &q_{\phi}(x^i_{t+s}=k|y^i_{t-\window+:t+h},w^i_{-t-\window+1:t}) = \beta^{i,k}_s\,,\\
        &\hat{\beta}^{i,k}_s = f_{\phi}(y^i_{t-\window+1:t},w^i_{t-\window+1:t})_{k,s}\,.
    \end{align*}
\end{itemize}
Architectures used for $(f^k_{\theta_y})_{1\leq k \leq K}, f_{\theta_x}$ and $f_{\phi}$ can be adjusted depending on the nature of the time series, the forecast horizon, if external signals are available, etc. Architectures used in the experiments section are detailed in Appendix~\ref{sec:fashiondatasetarchi} and Appendix~\ref{sec:referencedatasetarchi}. For completeness, a complete code base gathering the model implementation as well as the training process is publicly provided with this work\footnote{\url{https://github.com/etidav/next}}.

\section{Experiments}
\label{sec:nextresult}

In this section, we assess the performance of the proposed model on several datasets. The first experiment uses the dataset gathering 10000 fashion time series firstly introduced in \cite{david2022}. In this first experiment, the performance of our algorithm is evaluated and compared with several state-of-the-art methods. Moreover, as external signals are available, we show that the model can correctly leverage them to improve predictions. The proposed method is also evaluated with a collection of 8 reference datasets. This second application shows that the model can be easily applied to a wide variety of forecasting tasks and provide accurate predictions, rivaling complex state-of-the-art Transformer-based approaches.

\subsection{Fashion dataset}
\label{sec:nextresultfashiondataset}

\subsubsection{Fashion time series forecasting}

A first application of the proposed approach is done on the fashion dataset\footnote{\url{https://github.com/etidav/HERMES/}} introduced in \cite{david2022}. This dataset gathers a collection of 10000 weekly time series representing the evolution of the visibility of garments on social media. In addition, each sequence is linked with an external signal representing the visibility of the same garment on a sub sample of influencer users. The intuition is that influencers can adopt fashion items in advance and thus help forecasting methods to better predict the evolution of clothing on mainstream users. This dataset turned out to be well suited to our framework as it shows several specific features.
\begin{itemize}
    \item The fashion dataset contains numerous time series, showing thousands of different patterns of seasonality, trends, and noise levels. Some of the fashion time series are non-stationary.
    \item On some examples, early signals announcing the emergence of a new fashion item (which can also be considered as a change of regime) can be perceived in the external signals. Properly exploiting these additional signals could prove decisive in order to accurately detect and predict sudden changes present in the main time series.
\end{itemize}

\subsubsection{Baseline models and our model variants}

The following methods are tested on the fashion dataset as baseline approaches: \textit{Snaive}, \textit{Thetam} \citep{hyndman2020}, \textit{Ets} \citep{brown1961,Holt2004}, \textit{Tbats} \citep{alysha2011}, \textit{HERMES} \citep{david2022}, \textit{Prophet} \citep{Taylor2017}, \textit{N-BEATS} \citep{oreshkin2019}, \textit{N-HiTS} \citep{challu2023}, \textit{DeepAR} \citep{salinas2020} and \textit{PathTST} \citep{nie2023}. All these methods are reviewed in Appendix~\ref{sec:baselinemodelfashiondataset}. Against these methods, 2 variations of our approach with 2 hidden states are presented: i) a variation (mentioned as \textit{Ours}) that does not have access to the influencers external signals. ii) a variation having access to the external signals (mentioned as \textit{Ours-es} with '-es' for external signals). For the second method, external signals are included as input in the hidden state law and in only one of the two emission laws. Further information concerning parameters selection and the training process of the proposed approaches and some of the benchmark methods are reviewed in Appendix~\ref{sec:fashiondatasetgridsearch}.

\subsubsection{Accuracy metrics}

The fashion forecasting task is to predict the last year (52 values) of the 10000 time series. Evaluation of the tested methods accuracy is done using the Mean Absolute Scaled Error (MASE) as the fashion time series have different volumes:
\begin{align*}
\mathrm{MASE} &= \frac{T-m}{h}\frac{\sum_{j=1}^h |Y_{T+j} - \hat{Y}_{T+j}| }{\sum_{i=1}^{T-m} |Y_i - Y_{i-m}|}\,,
\end{align*}
where $T$ stand for the time series length, $h$ the forecast horizon and $m$ the seasonality (for the fashion dataset, $T=209$, $h=52$ and $m=52$). In addition of assessing the MASE on the whole dataset, the MASE is also evaluated on 2 sub samples of time series representing stationary and non-stationary time series. The following methodology is used to create these 2 samples.
\begin{itemize}
\item \textbf{non-stationary time series}. As a main challenge of the fashion forecasting use case is to correctly anticipate sudden evolution, a sub sample of time series showing strong non-stationary behaviours is studied. To create this sub sample of time series, the \textit{snaive} model is used to predict the last year of the fashion time series and the associated MASE are calculated. The non-stationary time series are defined as the 1000 time series where the \textit{snaive} prediction got the highest MASE.
\item \textbf{stationary time series}. By contrast, a group of stationary time series is presented. To define them, the same methodology as the previous group is used. We define them as the 1000 time series where the \textit{snaive} prediction reached the lowest MASE. 
\end{itemize}

\subsubsection{Results}

\begin{figure*}
\centering
  \includegraphics[width=.49\linewidth]{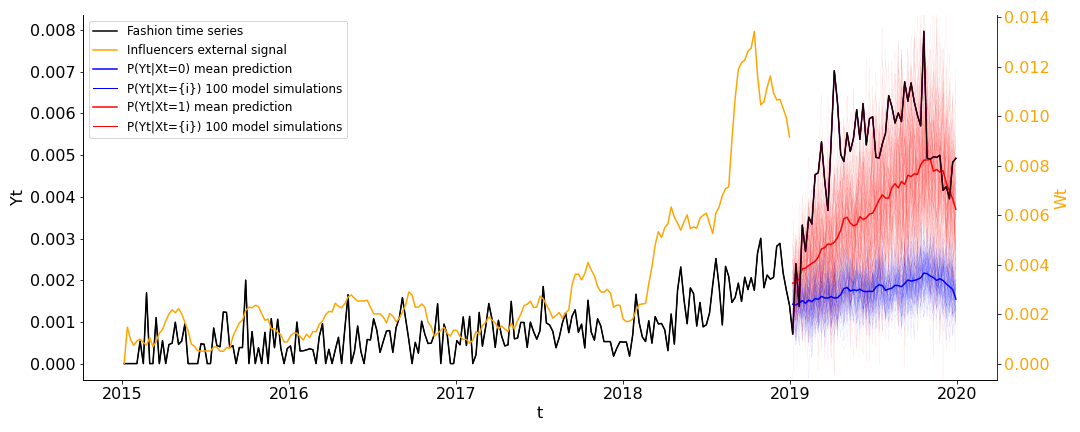}
  \includegraphics[width=.49\linewidth]{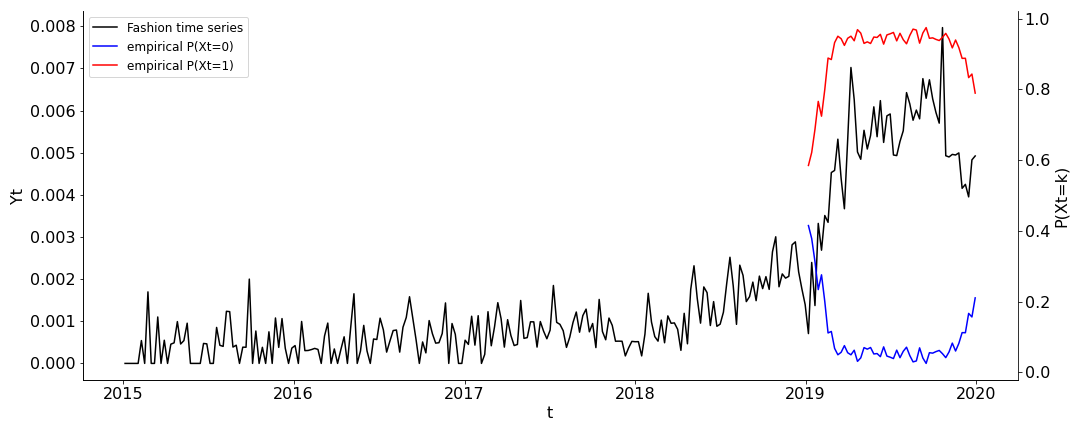}
  \includegraphics[width=.49\linewidth]{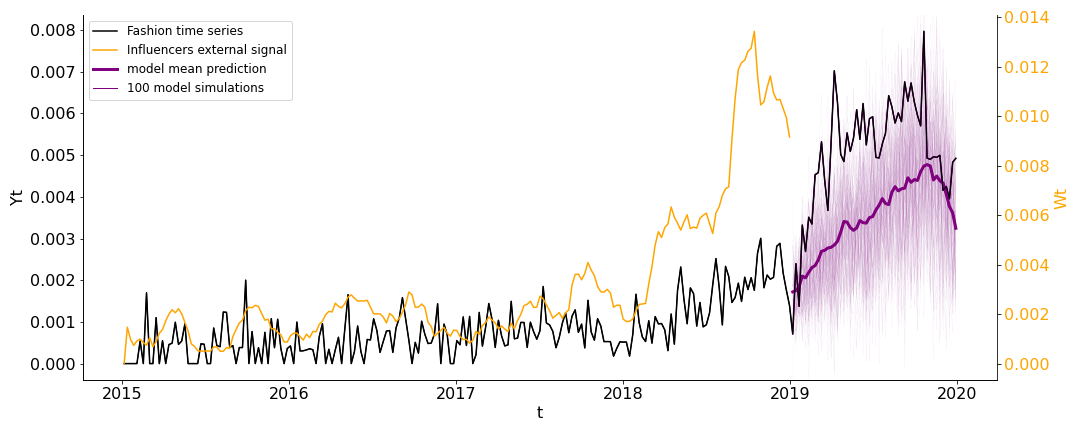}
\caption{\textbf{Predictions on the fashion dataset.} (Top Left) Prediction of the two emission laws when the hidden state is 0 or 1. (Top Right) Empirical distributions of the hidden states. (Bottom) Simulated predictions with our model using external signals.}
\label{fig:nextpredictionexample}
\end{figure*}

An example of model prediction on a fashion time series is displayed in Figure~\ref{fig:nextpredictionexample}. Hidden states trajectories, emission laws predictions and the final simulations are presented. In this example, the second emission law (emission law that has access to the external signal) catches the regime shift in the time series. This information is also correctly learnt as the empirical probability to be in this regime is close to one. Additional examples are provided in Appendix~\ref{sec:fashiondatasetpredictionexample}.

The final accuracy results are provided in Table~\ref{tab:nextdatasetaccuracy}. For each method, a prediction and the associated MASE are computed for the 10000 time series and the average is computed on the whole dataset, the non-stationary sample and the stationary sample. Among methods that have not access to the external signal, the proposed method (\textit{Ours}) has the highest accuracy on the whole dataset as well as on the 2 sub samples and outperforms other state-of-the-art models. The best results are provided by \textit{Ours-es}, the proposed method with the external signals. It outperforms all the other methods and shows a significant improvement, especially on the non-stationary time series.

\begin{table*}[t]
  \caption{\textbf{Fashion dataset accuracy results.} The Average MASE of each tested method is assessed on the whole dataset and 2 sub samples. For approaches using neural networks, 10 models are trained with different seeds. The mean and standard deviation of the 10 results computed with the 10 replicates are displayed.}
  \vspace{0.5cm}
  \centering
  \resizebox{\textwidth}{!}{
  \begin{tabular}{l||ccccc|ccccc|ccccc}
   &&\multicolumn{3}{c}{} &&& \multicolumn{3}{c}{\textbf{Non-stationary}} &&& \multicolumn{3}{c}{\textbf{Stationary}}&\\
   &&\multicolumn{3}{c}{\textbf{Fashion dataset}} &&& \multicolumn{3}{c}{\textbf{time series}} &&& \multicolumn{3}{c}{\textbf{time series}}&\\
    &&  \textit{MASE}  && \textit{seed std} &&&  \textit{MASE}  && \textit{seed std}&&&  \textit{MASE}  && \textit{seed std}& \\
	 \hline
	 &&&&&&&&&&\\
    \textit{Snaive} && 0.881 && - &&&1.455 && - &&& 0.536 && - \\
\textit{Thetam} && 0.844 && - &&&1.314 && - &&& 0.615 && - \\
\textit{Arima} && 0.826 && - &&&1.256 && - &&& 0.565 && - \\
\textit{Ets} && 0.807 && - &&&1.27 && - &&& 0.611 && - \\
\textit{Prophet} && 0.786 && - &&&1.193 && - &&& 0.629 && - \\
\textit{Stlm} && 0.77 && - &&&1.198 && - &&& 0.513 && - \\
\textit{Tbats} && 0.745 && - &&&1.229 && - &&& 0.501 && - \\
\textit{DeepAR} && 0.731 && 0.006 &&&1.158 && 0.031 &&& 0.508 && 0.017 \\
\textit{Hermes-ws} && 0.713 && 0.005 &&&1.092 && 0.007 &&& 0.477 && 0.008 \\
\textit{PatchTST} && 0.706 && 0.004 &&&1.149 && 0.01 &&& 0.448 && 0.003 \\
\textit{N-HiTS} && 0.701 && 0.003 &&&1.151 && 0.014 &&& 0.449 && 0.005 \\
\textit{N-BEATS} && 0.7 && 0.003 &&&1.146 && 0.014 &&& 0.451 && 0.003 \\
\textit{Ours} && 0.692 && 0.001 &&&1.116 && 0.006 &&& \textbf{0.44} && 0.001 \\
\textit{Ours-es} && \textbf{0.684} && 0.001 &&&\textbf{1.03} && 0.006 &&& 0.449 && 0.002 \\
  \end{tabular}}
\label{tab:nextdatasetaccuracy}
\end{table*} 

\subsubsection{Probabilistic forecast}

The proposed generative model allows to sample trajectories to assess the confidence of the forecast, see Figure~\ref{fig:nextpredictionexample} for an illustration. So as to evaluate the proposed approach on this specific point, 100 trajectories are computed with the method for each time series. Then, the MASE is computed for each trajectories and the average and standard deviation is displayed in Table~\ref{tab:probabilisticforecastaccuracymean}. The DeepAR method is also used as it allows to sample several predictions. We can see that the proposed model outperforms DeepAR and improves its probabilistic predictions when the influencers external signals are used.

\begin{table*}[t]
  \caption{\textbf{Fashion dataset probabilistic forecast accuracy results.} Final accuracy of methods providing probabilistic forecasts. For each method, 100 trajectories and their associated MASE are computed for each time series. The average and standard deviation is then calculated on the whole dataset and 2 sub samples.}
  \vspace{0.5cm}
  \centering
  \resizebox{\textwidth}{!}{
  \begin{tabular}{l||ccccc|ccccc|ccccc}
   &&\multicolumn{3}{c}{} &&& \multicolumn{3}{c}{\textbf{Non-stationary}} &&& \multicolumn{3}{c}{\textbf{Stationary}}&\\
   &&\multicolumn{3}{c}{\textbf{Fashion dataset}} &&& \multicolumn{3}{c}{\textbf{time series}} &&& \multicolumn{3}{c}{\textbf{time series}}&\\
   &&\multicolumn{3}{c}{\textit{MASE}} &&& \multicolumn{3}{c}{\textit{MASE}} &&& \multicolumn{3}{c}{\textit{MASE}}&\\
    &&  \textit{Mean}  && \textit{Std} &&&  \textit{Mean}  && \textit{Std}&&&  \textit{Mean}  && \textit{Std}& \\
	 \hline
	 &&&&&&&&&&\\
    \textit{DeepAR} && 0.969 && 0.339 &&&1.407 && 0.519 &&& 0.708 && 0.262  \\
    \textit{Ours} && 0.951 && 0.273 &&&1.364 && 0.394 &&& \textbf{0.655} && \textbf{0.153} \\
    \textit{Ours-es} && \textbf{0.943} && \textbf{0.268} &&&\textbf{1.319} && \textbf{0.362} &&& \textbf{0.656} && 0.166 \\
  \end{tabular}}
\label{tab:probabilisticforecastaccuracymean}
\end{table*}

\subsection{Reference dataset}
\label{sec:nextresultreferencedataset}

\subsubsection{Dataset presentation, Baseline models and the proposed approach}
The presented model is also evaluated with a collection of 8 reference datasets used in many recent contribution dealing with time series forecasting, see \cite{li2019,zhou2021,zhou2022,woo2022,wu2022,liu2022,zeng2022,wu2023,challu2023,woo2023,nie2023}. A review of these datasets is given in Appendix~\ref{sec:referencedatasetoverview}.

As benchmark against the proposed models, methods and results presented in the two following recent papers are used \cite{wu2023,nie2023}: the two best Transformer-based methods on the 8 datasets named PatchTST \citep{nie2023} and TimesNet \citep{wu2023}, a neural network called Dlinear that, as our approach, only relies and fully connected layers \citep{zeng2022} and 5 Transformer-based methods called FEDformer \citep{zhou2022}, Autoformer \citep{wu2022}, Informer \citep{zhou2021}, Pyraformer \citep{liu2022} and LogTrans \citep{li2019}.

Concerning the proposed approach, recurrent neural networks used on the fashion use case are replaced by fully connected networks as they are too computationally intensive for the long-term forecasting tasks (H=720). For all the reference datasets, the number of hidden states was set to 3 and the same architecture was used. Only a small grid search was run on each dataset for the shortest forecasting task to fix the length of the method inputs. Additional information concerning the proposed model on the reference dataset can be found in Appendix~\ref{sec:referencedatasetadditionalresult} and a code base is released to reproduce the results\footnote{\url{https://github.com/etidav/next}}.

\subsubsection{Accuracy metrics}
On the reference datasets, forecasting methods are evaluated on several horizons (lying between 24 to 720 time steps) and with 2 errors metrics, the Mean Square Error (MSE) and the Mean Absolute Error (MAE):
$$
\mathrm{MSE} = \frac{1}{h}\sum_{j=1}^h (Y_{T+j} - \hat{Y}_{T+j})^2\,, \qquad
\mathrm{MSE} = \frac{1}{h}\sum_{j=1}^h |Y_{T+j} - \hat{Y}_{T+j}|\,,
$$
with $h$ the forecast horizon. The last 20\% of each time series is kept hidden and used as test set.

\subsubsection{Results}

Table~\ref{tab:referencedatasetresult} displays the accuracy results of the benchmark models along with the proposed method on the 8 reference datasets. We can see that depending on the dataset and the horizon, the proposed method and the Transformer-based method PatchTST outperform all alternatives. These results illustrate two important features of the presented approach
\begin{itemize}
    \item The proposed method can be used for a large variety of time series forecasting task.
    \item By combining elementary neural networks components with hidden processes and a computationally efficient training procedure, the accuracy of our model reaches state-of-the-art standards and provide uncertainty quantification.
\end{itemize}
However, we can see that the Transformer-based model PatchTST outperforms the proposed model on some reference datasets such as Traffic or ECL. A main reason is that these two datasets gathers similar time series with long term evolution. In this context, the interest in introducing hidden states is low and the models used in the emission laws do not manage to outperform very complex and high dimensional models such as PatchTST. Additional numerical results on the reference datasets can be found in Appendix~\ref{sec:referencedatasetadditionalresult} along with examples of predictions for each reference dataset.

\renewcommand{\arraystretch}{1.2}
\begin{table*}[t]
  \caption{\textbf{Reference datasets accuracy results.} The best methods are highlithed in bold and the second best results with an underline.}
  \vspace{0.5cm}
  \centering
  \resizebox{\textwidth}{!}{
  \begin{tabular}{ll||llllllllllllllllllllll}
   && \multicolumn{2}{c}{Ours}&\multicolumn{2}{c}{PatchTST/64}&\multicolumn{2}{c}{TimesNet}&\multicolumn{2}{c}{DLinear}&\multicolumn{2}{c}{FEDformer}&\multicolumn{2}{c}{Autoformer}&\multicolumn{2}{c}{Informer}
   &\multicolumn{2}{c}{Pyraformer}
   &\multicolumn{2}{c}{LogTrans}\\
   &H&MSE&MAE&MSE&MAE&MSE&MAE&MSE&MAE&MSE&MAE&MSE&MAE&MSE&MAE&MSE&MAE&MSE&MAE\\
	 \hline
    \parbox[t]{2mm}{\multirow{4}{*}{\rotatebox[origin=c]{90}{\textit{Weather}}}}&96&\underline{0.154}&\underline{0.199}&\textbf{0.149}&\textbf{0.198}&0.172&0.220&0.176&0.237&0.217&0.296&0.266&0.336&0.300&0.384&0.896&0.556&0.458&0.490\\
&192&\underline{0.198}&\underline{0.242}&\textbf{0.194}&\textbf{0.241}&0.219&0.261&0.220&0.282&0.276&0.336&0.307&0.367&0.598&0.544&0.622&0.624&0.658&0.589\\
&336&\underline{0.252}&\underline{0.286}&\textbf{0.245}&\textbf{0.282}&0.280&0.306&0.265&0.319&0.339&0.38&0.359&0.395&0.578&0.523&0.739&0.753&0.797&0.652\\
&720&\underline{0.316}&\textbf{0.332}&\textbf{0.314}&\underline{0.334}&0.365&0.359&0.323&0.362&0.403&0.428&0.419&0.428&1.059&0.741&1.004&0.934&0.869&0.675\\
\hline
\parbox[t]{2mm}{\multirow{4}{*}{\rotatebox[origin=c]{90}{\textit{Traffic}}}}&96&\underline{0.396}&\underline{0.282}&\textbf{0.360}&\textbf{0.249}&0.593&0.321&0.410&\underline{0.282}&0.562&0.349&0.613&0.388&0.719&0.391&2.085&0.468&0.684&0.384\\
&192&\underline{0.423}&0.301&\textbf{0.379}&\textbf{0.256}&0.617&0.336&\underline{0.423}&\underline{0.287}&0.562&0.346&0.616&0.382&0.696&0.379&0.867&0.467&0.685&0.390\\
&336&0.437&0.306&\textbf{0.392}&\textbf{0.264}&0.629&0.336&\underline{0.436}&\underline{0.296}&0.570&0.323&0.622&0.337&0.777&0.420&0.869&0.469&0.733&0.408\\
&720&0.480&0.328&\textbf{0.432}&\textbf{0.286}&0.640&0.350&\underline{0.466}&\underline{0.315}&0.596&0.368&0.660&0.408&0.864&0.472&0.881&0.473&0.717&0.396\\
\hline
\parbox[t]{2mm}{\multirow{4}{*}{\rotatebox[origin=c]{90}{\textit{ECL}}}}&96&\underline{0.140}&0.240&\textbf{0.129}&\textbf{0.222}&0.168&0.272&\underline{0.140}&\underline{0.237}&0.183&0.297&0.201&0.317&0.274&0.368&0.386&0.449&0.258&0.357\\
&192&0.158&0.256&\textbf{0.147}&\textbf{0.240}&0.184&0.289&\underline{0.153}&\underline{0.249}&0.195&0.308&0.222&0.334&0.296&0.386&0.386&0.443&0.266&0.368\\
&336&0.176&0.274&\textbf{0.163}&\textbf{0.259}&0.198&0.300&\underline{0.169}&\underline{0.267}&0.212&0.313&0.231&0.338&0.300&0.394&0.378&0.443&0.280&0.380\\
&720&0.217&0.307&\textbf{0.197}&\textbf{0.290}&0.220&0.320&\underline{0.203}&\underline{0.301}&0.231&0.343&0.254&0.361&0.373&0.439&0.376&0.445&0.283&0.376\\
\hline
\parbox[t]{2mm}{\multirow{4}{*}{\rotatebox[origin=c]{90}{\textit{ILI}}}}&240&1.985&\underline{0.825}&\textbf{1.319}&\textbf{0.754}&2.317&0.934&2.215&1.081&2.203&0.963&3.483&1.287&5.764&1.677&\underline{1.420}&2.012&4.480&1.444\\
&36&\underline{1.746}&\textbf{0.783}&\textbf{1.579}&\underline{0.870}&1.972&0.920&1.963&0.963&2.272&0.976&3.103&1.148&4.755&1.467&7.394&2.031&4.799&1.467\\
&48&\underline{1.722}&\textbf{0.790}&\textbf{1.553}&\underline{0.815}&2.238&0.940&2.130&1.024&2.209&0.981&2.669&1.085&4.763&1.469&7.551&2.057&4.800&1.468\\
&60&\underline{1.684}&\underline{0.792}&\textbf{1.470}&\textbf{0.788}&2.027&0.928&2.368&1.096&2.545&1.061&2.770&1.125&5.264&1.564&7.662&2.100&5.278&1.560\\
\hline
\parbox[t]{2mm}{\multirow{4}{*}{\rotatebox[origin=c]{90}{\textit{ETTh1}}}}&96&0.379&\textbf{0.389}&\textbf{0.370}&0.400&0.384&0.402&\underline{0.375}&\underline{0.399}&0.376&0.419&0.449&0.459&0.865&0.713&0.664&0.612&0.878&0.740\\
&192&0.440&\underline{0.424}&\underline{0.413}&0.429&0.436&0.429&\textbf{0.405}&\textbf{0.416}&0.420&0.448&0.500&0.482&1.008&0.792&0.790&0.681&1.037&0.824\\
&336&0.483&0.445&\textbf{0.422}&\textbf{0.440}&0.491&0.469&\underline{0.439}&\underline{0.443}&0.459&0.465&0.521&0.496&1.107&0.809&0.891&0.738&1.238&0.932\\
&720&0.570&0.524&\textbf{0.447}&\textbf{0.468}&0.521&0.500&\underline{0.472}&\underline{0.490}&0.506&0.507&0.514&0.512&1.181&0.865&0.963&0.782&1.135&0.852\\
\hline
\parbox[t]{2mm}{\multirow{4}{*}{\rotatebox[origin=c]{90}{\textit{ETTh2}}}}&96&\textbf{0.271}&\textbf{0.332}&\underline{0.274}&\underline{0.337}&0.340&0.374&0.289&0.353&0.346&0.388&0.358&0.397&3.755&1.525&0.645&0.597&2.116&1.197\\
&192&\underline{0.347}&\textbf{0.382}&\textbf{0.341}&\textbf{0.382}&0.402&\underline{0.414}&0.383&0.418&0.429&0.439&0.456&0.452&5.602&1.931&0.788&0.683&4.315&1.635\\
&336&\underline{0.380}&\underline{0.409}&\textbf{0.329}&\textbf{0.384}&0.452&0.452&0.448&0.465&0.496&0.487&0.482&0.486&4.721&1.835&0.907&0.747&1.124&1.604\\
&720&\underline{0.420}&\underline{0.446}&\textbf{0.379}&\textbf{0.422}&0.462&0.468&0.605&0.551&0.463&0.474&0.515&0.511&3.647&1.625&0.963&0.783&3.188&1.540\\
\hline
\parbox[t]{2mm}{\multirow{4}{*}{\rotatebox[origin=c]{90}{\textit{ETTm1}}}}&96&\textbf{0.288}&\textbf{0.335}&\underline{0.293}&0.346&0.338&0.375&0.299&\underline{0.343}&0.379&0.419&0.505&0.475&0.672&0.571&0.543&0.510&0.600&0.546\\
&192&\textbf{0.331}&\textbf{0.363}&\underline{0.333}&0.370&0.374&0.387&0.335&\underline{0.365}&0.426&0.441&0.553&0.496&0.795&0.669&0.557&0.537&0.837&0.700\\
&336&\textbf{0.364}&\textbf{0.385}&\underline{0.369}&0.392&0.410&0.411&\underline{0.369}&\underline{0.386}&0.445&0.459&0.621&0.537&1.212&0.871&0.754&0.655&1.124&0.832\\
&720&0.429&0.426&\textbf{0.416}&\textbf{0.420}&0.478&0.450&\underline{0.425}&\underline{0.421}&0.543&0.490&0.671&0.561&1.166&0.823&0.908&0.724&1.153&0.820\\
\hline
\parbox[t]{2mm}{\multirow{4}{*}{\rotatebox[origin=c]{90}{\textit{ETTm2}}}}&96&\textbf{0.162}&\textbf{0.249}&\underline{0.166}&\underline{0.256}&0.187&0.267&0.167&0.260&0.203&0.287&0.255&0.339&0.365&0.453&0.435&0.507&0.768&0.642\\
&192&\textbf{0.218}&\textbf{0.288}&\underline{0.223}&\underline{0.296}&0.249&0.309&0.224&0.303&0.269&0.328&0.281&0.340&0.533&0.563&0.730&0.673&0.989&0.757\\
&336&\textbf{0.271}&\textbf{0.325}&\underline{0.274}&\underline{0.329}&0.321&0.351&0.281&0.342&0.325&0.366&0.339&0.372&1.363&0.887&1.201&0.845&1.334&0.872\\
&720&\textbf{0.355}&\textbf{0.380}&\underline{0.362}&\underline{0.385}&0.408&0.403&0.397&0.421&0.421&0.415&0.422&0.419&3.379&1.388&3.625&1.451&3.048&1.328\\
\hline
  \end{tabular}}
\label{tab:referencedatasetresult}
\end{table*}
\renewcommand{\arraystretch}{1.}

\section{Conclusion}
\label{sec:nextconclusion}

In this paper, a new time series forecasting model combining discrete hidden Markov models and deep architectures is introduced. Depending on the dynamics of the hidden states, several emission laws are learned and activated to produce the final predictions. We proposed a two-stage training procedure, based on the ELBO and inspired by recent variational quantization approaches. Our model was first tested on a fashion dataset and outperformed state-of-the-art methods in particular when using external signals. Then, its performance were assessed on 8 reference datasets with similar performance as Transformer-based methods.\\
The approach allows to obtain an estimation of the predictive distribution of future observations. This generative model can be further investigated by providing an automatic selection of the number hidden states, and extending recent results on variational learning of hidden Markov models to obtain theoretical guarantees on the variational distribution.

\bibliography{next_paper}
\bibliographystyle{apalike}

\appendix
\section{Fashion dataset}

\subsection{Benchmark models}
\label{sec:baselinemodelfashiondataset}

We present in this section the baseline methods tested on the fashion dataset against the proposed approach:
\begin{itemize}
    \item \textbf{Snaive:} A method that only repeats the last past period of historical data.
    \item \textbf{Thetam:} A parametric model that decomposes the original signal in $\theta$-lines, predicts each one separately and recomposes them to produce the final forecast (\cite{hyndman2020}).
    \item \textbf{Ets:} The exponential smoothing method (\cite{brown1961},\cite{Holt2004}).
    \item \textbf{Tbats:} A parametric model presented in \cite{alysha2011}.
    \item \textbf{Stlm:} A parametric model that uses a multiplicative decomposition and models the seasonally adjusted time series with an exponential smoothing model (\cite{hyndman2020}).
    \item \textbf{HERMES:} a hybrid method mixing per-time-series TBATS predictors and a recurent neural network global corrector \cite{david2022}.
    \item \textbf{Prophet:} a parametric model introduced in \cite{Taylor2017} and widely used in the indutry.
    \item \textbf{N-BEATS:} a full-neural-network-based method that shows striking results on numerous datasets of the literature \cite{oreshkin2019}.
    \item \textbf{N-HiTS:} The evolution of N-BEATS \cite{challu2023}.
    \item \textbf{DeepAR:} a full-neural-network-based method used at Amazon that provided sharp probabilistic forecasts \cite{salinas2020}.
    \item \textbf{PatchTST:} A Transformer-based model that emerged as the best method using Transformers on several datasets of the literature \cite{nie2023}.
\end{itemize}
For the methods DeepAR, N-BEATS, N-HiTS and PatchTST, the package "neuralforecast" was used to train them \citep{olivares2022}.

\subsection{Architecture used for hidden states and emission laws}
\label{sec:fashiondatasetarchi}

\begin{figure}
    \centering
    \includegraphics[width=1\linewidth]{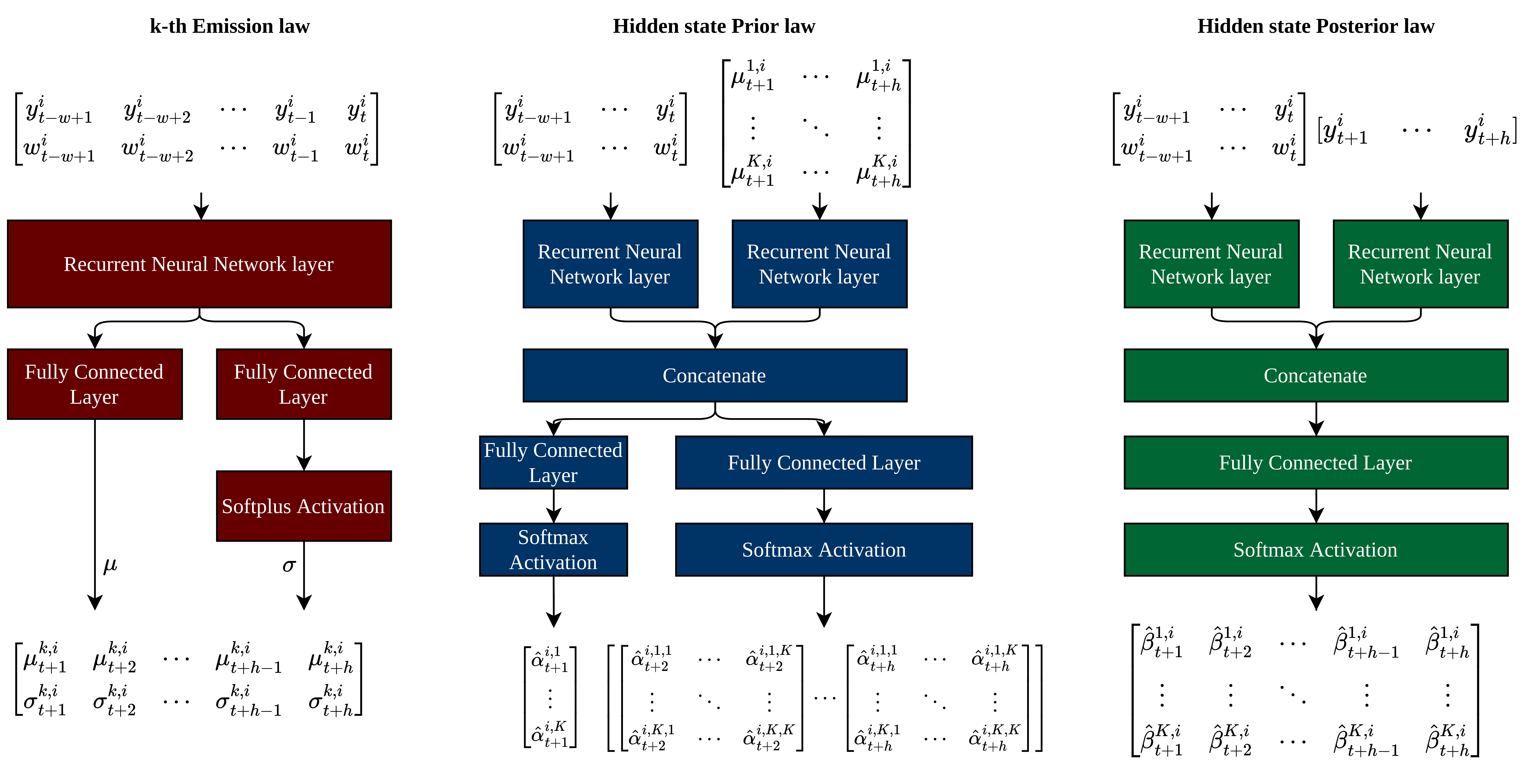}
    \caption{\textbf{Example of model architecture.} Example of architecture used for the proposed approach on the Fashion dataset. (Left) Model used to compute parameters of the k-th emission law. (Middle) Model used to compute the hidden state probabilities. (Right) Model used to approximate the posterior law of the hidden states.}
    \label{fig:nextarchi}
\end{figure}

We detail in this section the architecture used for the proposed model on the Fashion dataset. For the emission laws, a LSTM layer is first used to process the past of the main signal and external signals. Then, Fully Connected (FC) layers are used to compute the different parameters of the emission laws. For the standard deviation of the Gaussian emission laws, a "softplus" activation is applied on the last layer to ensure that the model outputs remain positive. Concerning the hidden state prior law, two LSTM layers are used to process the past inputs (main signals plus external signals) and some outputs of the emission laws. Outputs are concatenated and fed to two Fully Connected layers followed by a "softmax" activation to compute the initial law and transition matrices of the hidden state processes. Finally, for the posterior law of the hidden states, past and future windows of the main signal are first provided to two LSTM layers. Outputs are concatenate and fed to a Fully Connected layers and a "softmax" activation to compute the posterior variational probabilities. See Figure~\ref{fig:nextarchi} for an illustration of the different components.

\subsection{Fixing the number of hidden states}

Table~\ref{tab:nextdatasetaccuracyhs} displays results of several variations of the proposed model with a number of hidden states between 2 and 4. We see that the variation achieving the best accuracy is the method with two hidden states and that increasing the number of hidden states does not always lead to an increase in final accuracy. Indeed, we notice that the more the model has hidden states, the more difficult it is to differentiate them, which leads to redundant emission laws. See Appendix~\ref{sec:fashiondatasetpredictionexample} for prediction examples of the variation with 4 hidden states. Future works will focus on providing an automatic selection of $K$. %Further work will be done to understand how it could be possible to help the proposed model to easier handle more hidden states.

\begin{table*}
  \caption{\textbf{Hidden states grid search.} Average MASE of the proposed model with a number of hidden states lying between 2 and 4 are assessed on the Fashion dataset. For each model, 10 models are trained with different seeds. The mean and standard deviation of the 10 results computed with the 10 replicates are displayed.}
  \vspace{0.5cm}
  \centering
  \resizebox{\textwidth}{!}{
  \begin{tabular}{l||ccccc|ccccc|ccccc}
   &&\multicolumn{3}{c}{} &&& \multicolumn{3}{c}{\textbf{Non-stationary}} &&& \multicolumn{3}{c}{\textbf{Stationary}}&\\
   &&\multicolumn{3}{c}{\textbf{Fashion dataset}} &&& \multicolumn{3}{c}{\textbf{time series}} &&& \multicolumn{3}{c}{\textbf{time series}}&\\
    &&  \textit{MASE}  && \textit{seed std} &&&  \textit{MASE}  && \textit{seed std}&&&  \textit{MASE}  && \textit{seed std}& \\
	 \hline
	 &&&&&&&&&&\\
    \textit{Ours 3hs} && 0.693 && 0.001 &&&1.118 && 0.006 &&& 0.441 && 0.002 \\
    \textit{Ours 4hs} && 0.693 && 0.001 &&&1.113 && 0.004 &&& 0.442 && 0.001 \\
    \textit{Ours 2hs} && 0.692 && 0.001 &&&1.116 && 0.006 &&& \textbf{0.44} && 0.001 \\
    \textit{Ours-es 3hs} && 0.685 && 0.001 &&&1.031 && 0.005 &&& 0.452 && 0.002 \\
    \textit{Ours-es 4hs} && 0.685 && 0.001 &&&\textbf{1.029} && 0.005 &&& 0.452 && 0.002 \\
    \textit{Ours-es 2hs} && \textbf{0.684} && 0.001 &&&1.03 && 0.006 &&& 0.449 && 0.002 \\
  \end{tabular}}
\label{tab:nextdatasetaccuracyhs}
\end{table*} 

\subsection{Grid search}
\label{sec:fashiondatasetgridsearch}

So as to produce the final results of the benchmark methods and the proposed model on the Fashion dataset, several grid searches were run to fix the different hyper parameters. For the methods PatchTST, N-HiTS, N-BEATS, DeepAR and our method, a grid search was run on the learning rate and the batch size. Table~\ref{tab:gridsearchbenchmarkfashiondataset} summarizes the grid search results for these 5 models. The best configuration in terms of MASE on the test set was selected and used to produce the final results displayed in Section~\ref{sec:nextresultfashiondataset}.

\begin{table}
  \caption{\textbf{Fashion dataset benchmarks grid search} Grid searches run on the Fashion dataset for the following benchmark methods: DeepAR, PatchTST, N-HiTS, N-BEATS and our method. The metrics displayed are the final MASE of each model variation on the test set.}
  \vspace{0.5cm} 
 \centering
 \resizebox{1\textwidth}{!}{
 \begin{tabular}{ll||lllll}
 \multicolumn{5}{c}{\textit{DeepAR}}\\
   &&\multicolumn{3}{c}{\textbf{Learning rate}}\\
    && \textit{0.05} & \textit{0.005} & \textit{0.0005} \\
	 \hline
	 &&&&\\
    \parbox[t]{2mm}{\multirow{5}{*}{\rotatebox[origin=c]{90}{\textbf{Batch size}}}}&8&0.76&0.791&0.874\\
    &64&0.733&0.736&0.831\\
    &256&0.774&0.771&0.772\\
    &1024&0.754&0.75&0.754\\
    &2048&\textbf{0.727}&0.745&0.752\\
  \end{tabular}\hspace{1cm}
  \begin{tabular}{ll||lllll}
  \multicolumn{5}{c}{\textit{PatchTST}}\\
   &&\multicolumn{3}{c}{\textbf{Learning rate}}\\
    && \textit{0.05} & \textit{0.005} & \textit{0.0005} \\
	 \hline
	 &&&&\\
    \parbox[t]{2mm}{\multirow{5}{*}{\rotatebox[origin=c]{90}{\textbf{Batch size}}}}&8&0.85&0.714&0.717\\
    &64&0.881&0.705&0.707\\
    &256&0.913&0.705&0.708\\
    &1024&0.818&\textbf{0.704}&0.709\\
    &2048&0.947&0.709&0.709\\
  \end{tabular}\hspace{1cm}
  \begin{tabular}{ll||lllll}
  \multicolumn{5}{c}{\textit{Ours}}\\
   &&\multicolumn{3}{c}{\textbf{Learning rate}}\\
    && \textit{0.005} & \textit{0.0005} & \textit{0.00005} \\
	 \hline
	 &&&&\\
    \parbox[t]{2mm}{\multirow{4}{*}{\rotatebox[origin=c]{90}{\textbf{Batch size}}}}&64&0.714&0.713&0.727\\
    &256&0.703&0.700&0.714\\
    &1024&0.694&0.696&0.709\\
    &2048&0.694&\textbf{0.693}&0.702\\
  \end{tabular}
 }\vspace{1cm}
 \resizebox{0.66\textwidth}{!}{ 
  \begin{tabular}{ll||lllll}
  \multicolumn{5}{c}{\textit{N-HiTS}}\\
   &&\multicolumn{3}{c}{\textbf{Learning rate}}\\
    && \textit{0.05} & \textit{0.005} & \textit{0.0005} \\
	 \hline
	 &&&&\\
    \parbox[t]{2mm}{\multirow{5}{*}{\rotatebox[origin=c]{90}{\textbf{Batch size}}}}&8&0.733&0.709&\textbf{0.701}\\
    &64&0.716&0.733&0.702\\
    &256&0.719&0.733&0.702\\
    &1024&0.715&0.734&0.702\\
    &2048&0.717&0.734&0.703\\
  \end{tabular}\hspace{1cm}
  \begin{tabular}{ll||lllll}
  \multicolumn{5}{c}{\textit{N-BEATS}}\\
   &&\multicolumn{3}{c}{\textbf{Learning rate}}\\
    && \textit{0.05} & \textit{0.005} & \textit{0.0005} \\
	 \hline
	 &&&&\\
    \parbox[t]{2mm}{\multirow{5}{*}{\rotatebox[origin=c]{90}{\textbf{Batch size}}}}&8&0.740&0.709&\textbf{0.700}\\
    &64&0.713&0.734&0.702\\
    &256&0.719&0.738&0.703\\
    &1024&0.718&0.737&0.704\\
    &2048&0.876&0.741&0.704\\
  \end{tabular}
  }
 \label{tab:gridsearchbenchmarkfashiondataset}
\end{table}

\subsection{Example of predictions}
\label{sec:fashiondatasetpredictionexample}

\begin{figure*}
\centering
  \includegraphics[width=1.\linewidth]{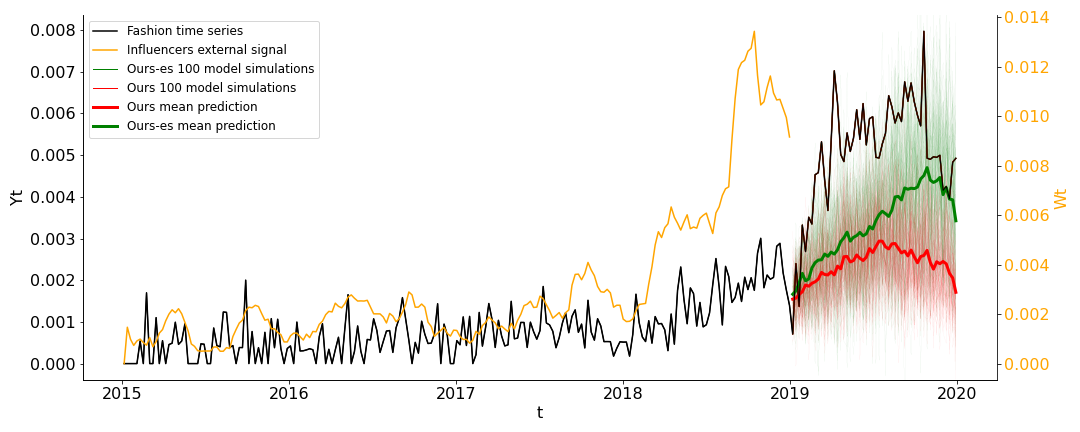}
  \includegraphics[width=1.\linewidth]{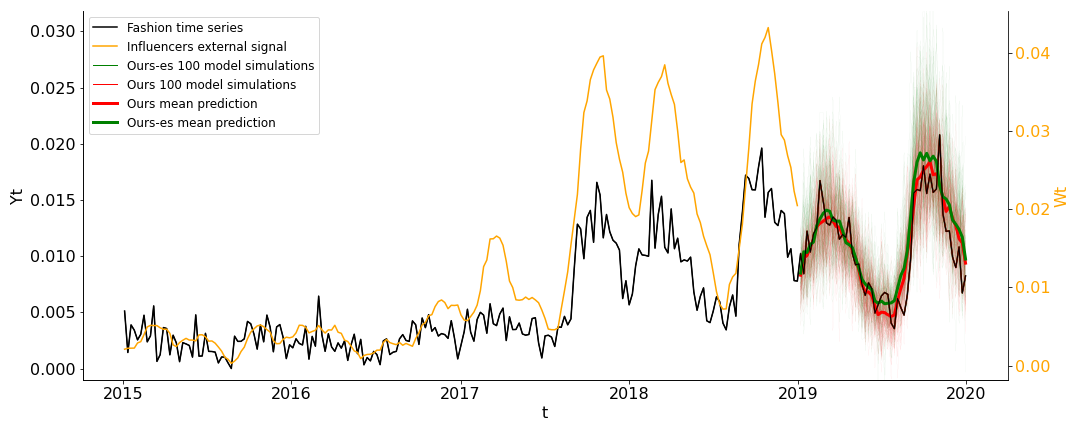}
  \includegraphics[width=1.\linewidth]{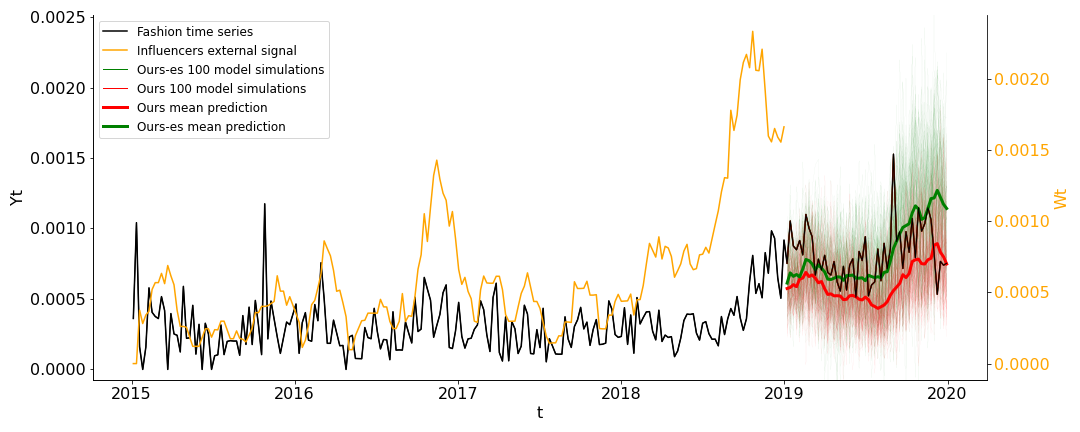}
\caption{\textbf{Ours vs Ours-es predictions.} \textit{Ours} and \textit{Ours-es} model predictions on three fashion time series: (Top) "br\_female\_shoes\_262", (Middle) "eu\_female\_outerwear\_177", (Bottom) "eu\_female\_texture\_80". On several fashion time series, \textit{Ours-es} correctly leverages the influencers external signal and capture sudden non-stationary evolution impossible to forecast without them.}
\label{fig:nextnextffpredictionexample}
\end{figure*}

\begin{figure*}
\centering
  \includegraphics[width=1.\linewidth]{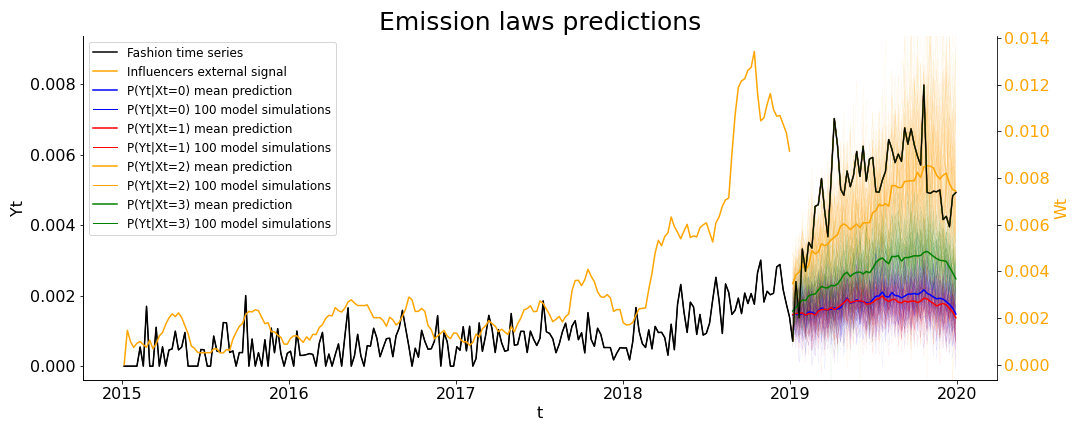}
  \includegraphics[width=1.\linewidth]{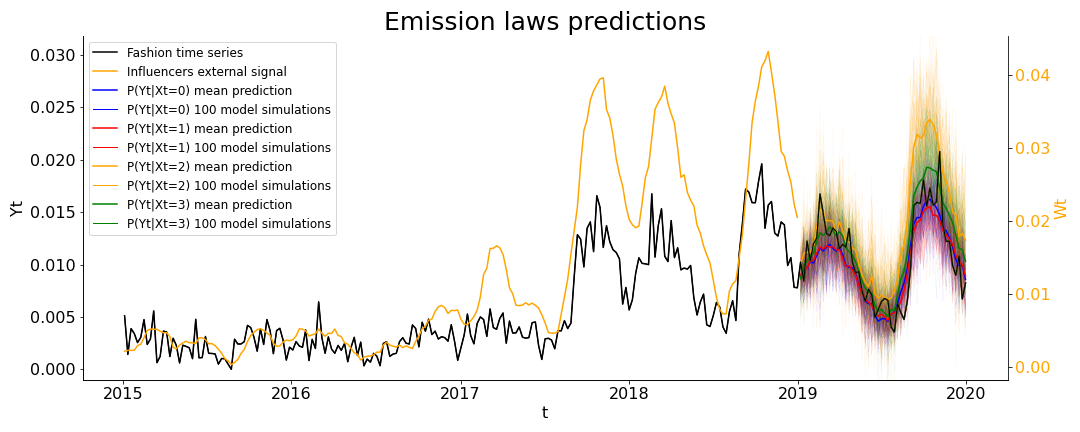}
  \includegraphics[width=1.\linewidth]{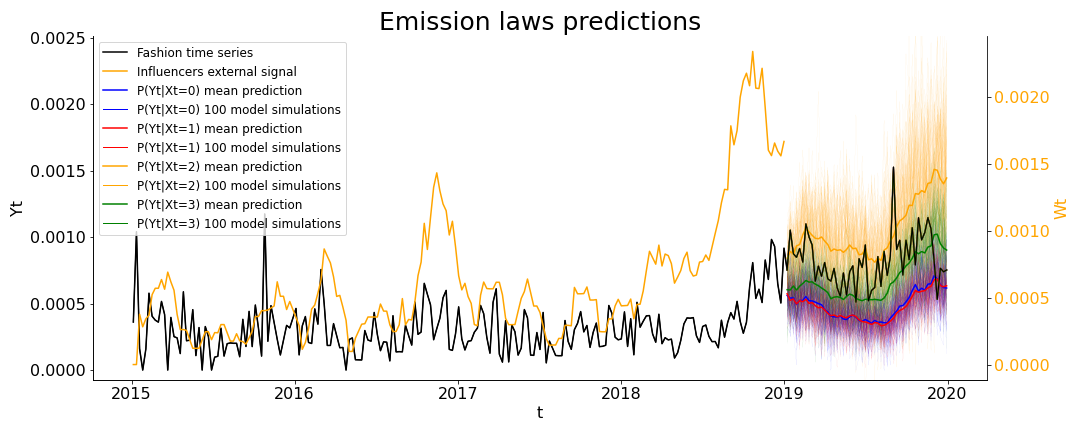}
\caption{\textbf{Proposed method with 4 hidden states predictions.} Emission laws predictions of the proposed model with 4 hidden states on three fashion time series: (Top) "br\_female\_shoes\_262", (Middle) "eu\_female\_outerwear\_177", (Bottom) "eu\_female\_texture\_80". For this model, the influencers external signal was only given to the third and the fourth emission laws. The third and the fourth emission laws learned different distribution but the first and the second ones seems to be redundant.}
\label{fig:4hspredictionexample}
\end{figure*}

\begin{figure*}
\centering
  \includegraphics[width=1.\linewidth]{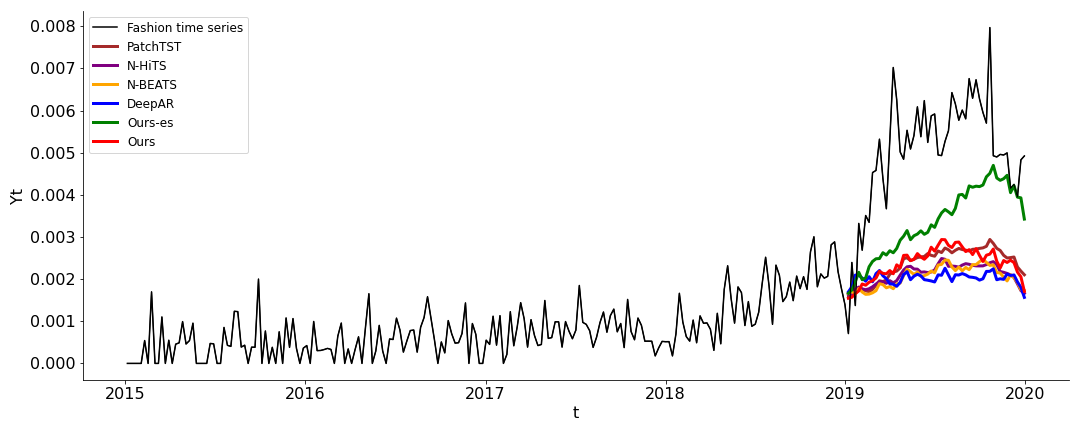}
  \includegraphics[width=1.\linewidth]{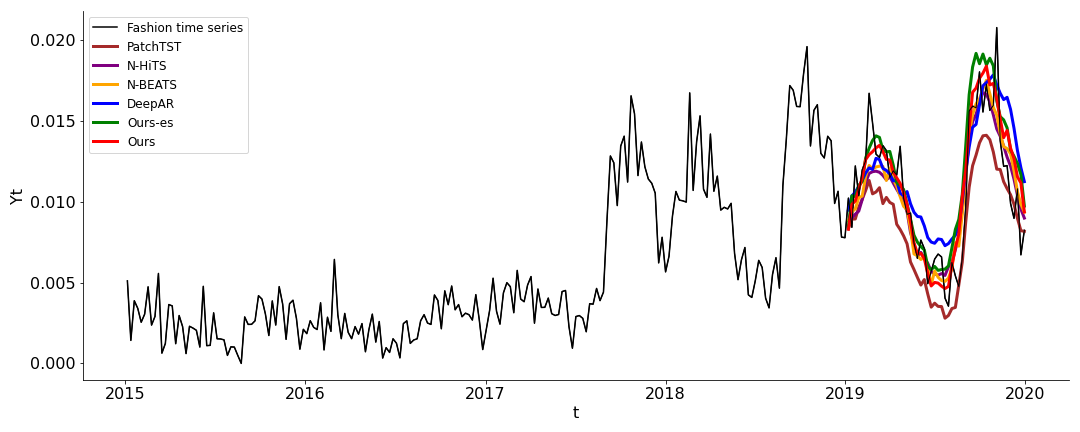}
  \includegraphics[width=1.\linewidth]{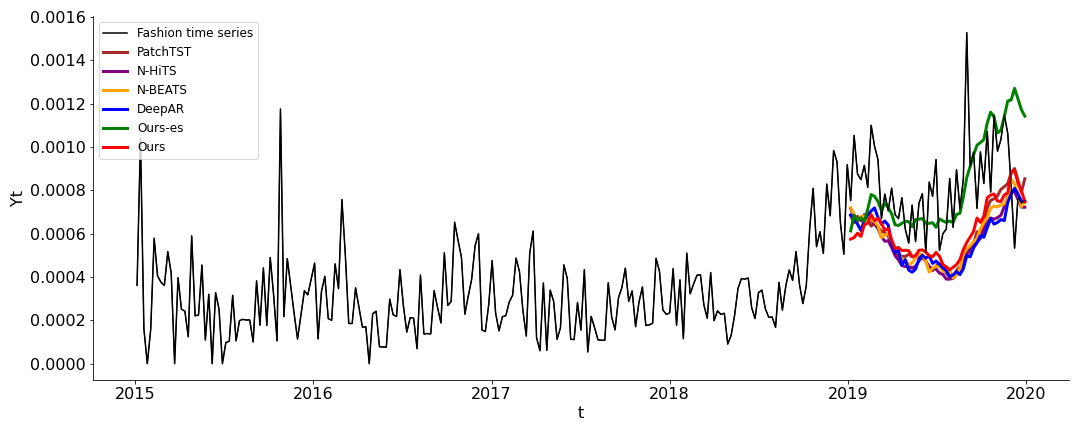}
\caption{\textbf{Presented method and benchmark models predictions.} Final prediction of the presented model and some of the benchmark methods on three fashion time series (Top) "br\_female\_shoes\_262", (Middle) "eu\_female\_outerwear\_177", (Bottom) "eu\_female\_texture\_80". The model \textit{Ours} seems to compute more accurate predictions than benchmark methods on these examples but the best forecast are provided by the model \textit{Ours-es} with the use of the influencers external signal.}
\label{fig:allpredictionexample}
\end{figure*}

Finally, we display additional examples of predictions using the proposed model on the Fashion dataset. First, Figure~\ref{fig:nextnextffpredictionexample} displays a comparison between predictions of the model having access to the influencers external signals and without having access to them. We can see that in some examples, the inclusion of the external signals greatly helps one of the emission laws to explore new distributions and accurately catch non-stationary evolution. Then, Figure~\ref{fig:4hspredictionexample} shows the prediction of the presented model with 4 hidden states and illustrates that adding hidden states does not necessarily lead to a better exploration of the dataset but may lead to redundant regimes. Finally, Figure~\ref{fig:allpredictionexample} displays examples of prediction of the proposed model along with some of the best benchmark models.

\section{Reference dataset}
\label{sec:referencedatasetadditionalresult}

\subsection{Reference Datasets}
\label{sec:referencedatasetoverview}

We present in this section the 8 references dataset used in Section~\ref{sec:nextresultreferencedataset}.
\begin{itemize}
    \item \textbf{ETTm2} (Electricity Transformer Temperature): a dataset gathering time series following characteristics of an electricity transformer in China from July 2016 to July 2018 with values measured every 15 minutes \cite{haoyietal2021}.
    \item \textbf{ECL}(Electricity): time series representing the evolution of the electricity consumption of 370 clients from 2012 to 2014 \cite{trindade2015}.
    \item \textbf{Exchange-Rate}: a dataset gathering 8 time series representing the evolution from 1990 to 2016 of the daily exchange rates of the following countries: Australia, British, Canada, Switzerland, China, Japan, New Zealand and Singapore \cite{lai2018}.
    \item \textbf{Traffic} (San Francisco Bay Area Highway Traffic): 862 time series representing roads occupancy measured by 862 sensors spread over the State of California from January 2015 to December 2016.
    \item \textbf{Weather}: dataset gathering the evolution of 21 meteorological variables in Germany during the year 2020.
    \item \textbf{ILI}(Influenza-like illness): time series representing the weekly evolution of the number of influenza-like illness patient in The United States, from January 2002 to July 2020.
\end{itemize}

\subsection{Architecture used for hidden states and emission laws}
\label{sec:referencedatasetarchi}

\begin{figure}
    \centering
    \includegraphics[width=1\linewidth]{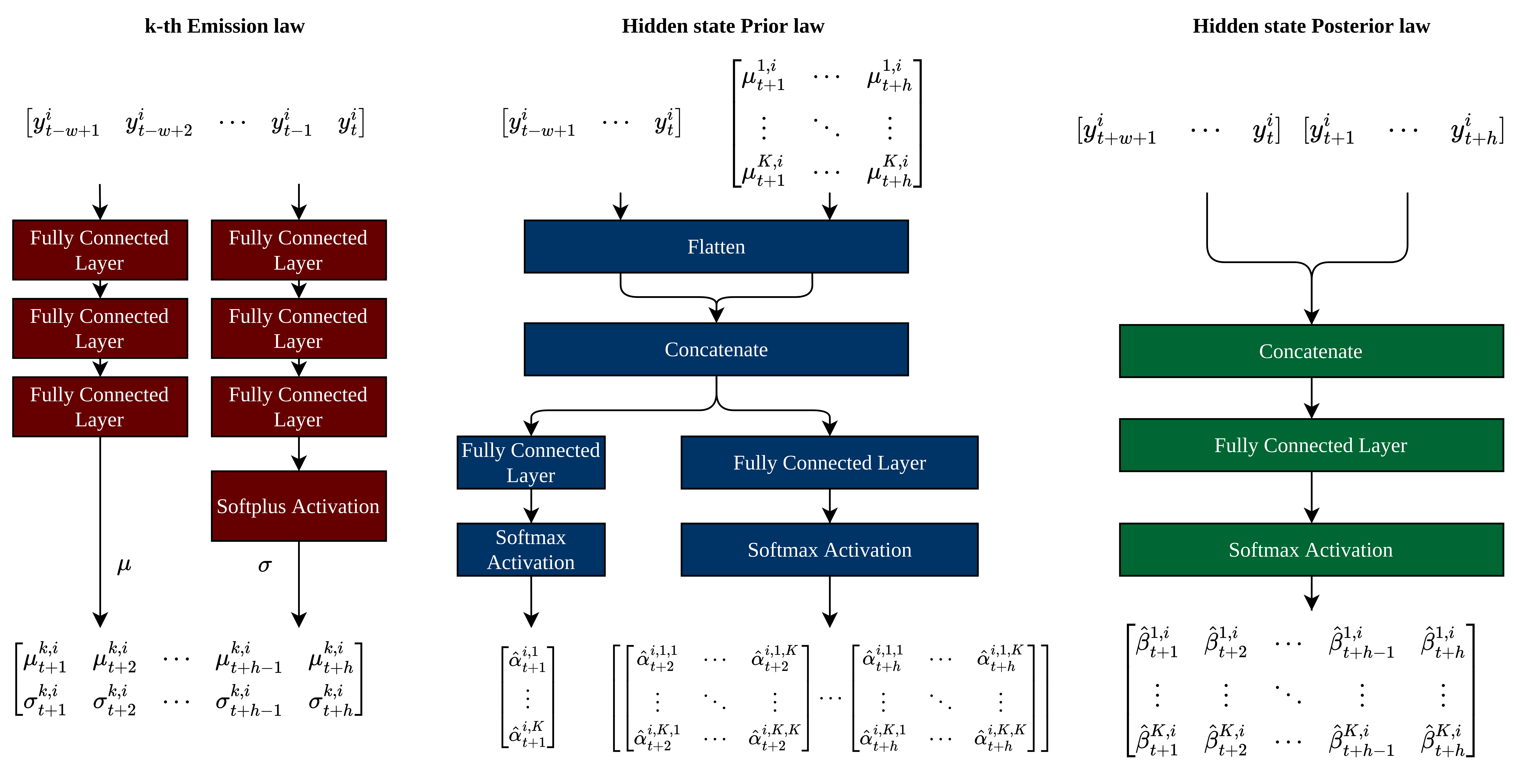}
    \caption{\textbf{Example of model architecture.} Example of architecture used for the proposed approach on the 8 reference datasets. (Left) Model used to compute parameters of the k-th emission law. (Middle) Model used to compute the hidden state probabilities. (Right) Model used to approximate the posterior variational law of the hidden states.}
    \label{fig:referencedatasetnextarchi}
\end{figure}

An overview of the architecture used for the proposed model on the 8 reference dataset is displayed in Figure~\ref{fig:referencedatasetnextarchi}. As recurrent neural layers considerably slow down the model for some of the long-term forecasting tasks (especially where horizon=720), they are all replaced by fully connected layers. Except for this modification, the architecture used with the reference datasets is similar to that used on the Fashion dataset.

\subsection{The importance of hidden states}

On all the 8 benchmark datasets, we fix the number of hidden states to 3 for the proposed model. However, as for the past dependency parameter, a gridsearch can be done to find the optimal number of hidden states. On three reference datasets (Traffic, Weather and ETTh2) and for the forecasting task where horizon is fixed to 96, we train 4 variations of the proposed method with a number of hidden states lying between 1 to 4. Results are displayed in Figure~\ref{tab:hsgridsearchnextrefdataset}. We can see that the optimal number of hidden states can change depending on the use case but the difference in terms of accuracy remains low between 2 and 4 hidden states.

\begin{table}
  \caption{\textbf{Hidden state parameter} Analysis of the importance of the number of hidden states on 3 of the 8 reference datasets. We test a number of hidden states from 1 to 4. The metrics displayed are the final MSE and MAE on the validation and test set.}
  \vspace{0.5cm} 
  \centering
  \resizebox{\textwidth}{!}{
  \begin{tabular}{ll||llllllllllllllll}
   \multicolumn{2}{c||}{\textit{hidden states}}& \multicolumn{4}{c}{1 hidden state}&\multicolumn{4}{c}{2 hidden states}& \multicolumn{4}{c}{3 hidden states}&\multicolumn{4}{c}{4 hidden states}\\
   && \multicolumn{2}{c}{Eval}&\multicolumn{2}{c}{Test}&\multicolumn{2}{c}{Eval}&\multicolumn{2}{c}{Test}&\multicolumn{2}{c}{Eval}&\multicolumn{2}{c}{Test}&\multicolumn{2}{c}{Eval}&\multicolumn{2}{c}{Test}\\
   dataset&H&MSE&MAE&MSE&MAE&MSE&MAE&MSE&MAE&MSE&MAE&MSE&MAE&MSE&MAE&MSE&MAE\\
	 \hline
    Weather&96&0.399&0.280&0.155&0.201&0.397&0.278&0.154&0.200&0.398&0.276&\textbf{0.153}&\textbf{0.199}&0.399&0.277&0.154&0.200\\
    Traffic&96&0.329&0.244&\textbf{0.399}&0.287&0.330&0.243&0.400&0.286&0.330&0.242&\textbf{0.399}&\textbf{0.284}&0.330&0.242&\textbf{0.399}&0.285\\
    ETTh2&96&0.236&0.316&0.273&0.334&0.239&0.313&0.276&0.332&0.238&0.311&0.273&\textbf{0.331}&0.236&0.310&\textbf{0.272}&\textbf{0.331}\\
  \end{tabular}}
 \label{tab:hsgridsearchnextrefdataset}
\end{table}

\subsection{Minmaxscaler versus Standardscaler}

On the 8 reference datasets, we investigate the potential impact of the preprocessing step on the proposed model. Consequently, on 4 of the 8 reference datasets (ETTh1, ETTh2, ETTm1 and ETTm2) and for the forecasting task where horizon is fixed to 96, the two normalization included by the proposed model (Minmaxscaler and StandardScaler) are tested. Table~\ref{tab:gridsearchnormnextrefdataset} displays results of the different trainings and we can see that accuracy results can be strongly impacted by the preprocessing. As the Minmaxscaler normalization seems to be more robust than the Standardscaler, it was selected for the proposed architecture on the 8 reference datasets.

\begin{table}
  \caption{\textbf{Preprocessing analysis} Analysis of the impact of the preprocessing on the proposed method. The Minmaxscaler (scale the inputs between 0 and 1) and the Standardscaler (scale the inputs to have mean 0 and a variance of 1) approach are tested. The metrics displayed are the final MSE and MAE on the validation and test set.}
  \vspace{0.5cm} 
  \centering
  \resizebox{\textwidth}{!}{
  \begin{tabular}{ll||llllllllllllllll}
   \multicolumn{2}{c||}{\textit{preprocess name}}& \multicolumn{4}{c}{MinMaxscaler}&\multicolumn{4}{c}{Standardscaler}\\
   && \multicolumn{2}{c}{Eval}&\multicolumn{2}{c}{Test}&\multicolumn{2}{c}{Eval}&\multicolumn{2}{c}{Test}\\
   dataset&H&MSE&MAE&MSE&MAE&MSE&MAE&MSE&MAE\\
	 \hline
    ETTh1&96&\textbf{0.487}&\textbf{0.451}&0.379&0.389&0.494&0.458&0.380&0.394\\
    ETTh2&96&\textbf{0.239}&\textbf{0.315}&0.271&0.332&0.271&0.348&0.306&0.362\\
    ETTm1&96&\textbf{0.303}&\textbf{0.354}&0.288&0.336&0.304&0.355&0.292&0.340\\
    ETTm2&96&\textbf{0.124}&\textbf{0.233}&0.162&0.249&0.134&0.241&0.167&0.253\\
  \end{tabular}}
 \label{tab:gridsearchnormnextrefdataset}
\end{table}

\subsection{past dependency grid search}

On the 8 reference datasets presented in Section~\ref{sec:nextresultreferencedataset}, a gridsearch was run to set the best past dependency length for the proposed approach. For each dataset, several input sizes were tested from half of the seasonally to 8 times the seasonality. The best one was selected based on the MSE of the resulting model on the validation set. Table~\ref{tab:gridsearchnextrefdataset} summarizes the results of each gridsearch.

\begin{table}
  \caption{\textbf{Past dependency grid search} Grid searches run on the 8 reference datasets to fix the optimal past dependency parameter for the proposed model. We tested a range of values between the half of the seasonally to 8 times the seasonality. The metrics displayed are the final MSE and MAE on the validation set.}
  \vspace{0.5cm} 
  \centering
  \resizebox{\textwidth}{!}{
  \begin{tabular}{ll||llllllllllllllll}
   \multicolumn{2}{c||}{\textit{Past dependency}}& \multicolumn{2}{c}{0.5*seasonality}&\multicolumn{2}{c}{1*seasonality}&\multicolumn{2}{c}{2*seasonality}&\multicolumn{2}{c}{3*seasonality}&\multicolumn{2}{c}{4*seasonality}&\multicolumn{2}{c}{5*seasonality}\\
   dataset&H&MSE&MAE&MSE&MAE&MSE&MAE&MSE&MAE&MSE&MAE&MSE&MAE\\
	 \hline
    Weather&96&0.639&0.346&0.470&0.290&0.433&0.277&0.423&\textbf{0.276}&0.409&\textbf{0.276}&\textbf{0.400}&\textbf{0.276}\\
    Traffic&96&0.537&0.333&0.366&0.254&0.337&0.243&0.329&0.241&0.328&0.240&\textbf{0.324}&\textbf{0.239}\\
    ECL&96&0.183&0.256&0.132&0.225&0.122&0.217&0.119&\textbf{0.215}&\textbf{0.118}&\textbf{0.215}&\textbf{0.118}&0.216\\
    ILI&96&0.322&0.405&\textbf{0.146}&\textbf{0.232}&0.296&0.317&0.237&0.276&0.217&0.314&0.263&0.373\\
    ETTh1&96&\textbf{0.485}&\textbf{0.450}&0.500&0.465&0.498&0.476&0.512&0.490&0.512&0.490&0.520&0.499\\
    ETTh2&96&0.244&0.319&\textbf{0.234}&\textbf{0.309}&0.239&0.312&0.239&0.312&0.237&0.313&0.242&0.320\\
    ETTm1&96&0.463&0.439&0.345&0.375&0.314&0.360&0.309&0.360&\textbf{0.306}&\textbf{0.356}&0.308&0.357\\
    ETTm2&96&0.141&0.250&0.134&0.242&0.131&0.239&\textbf{0.125}&\textbf{0.234}&0.129&0.237&0.126&\textbf{0.234}\\
  \end{tabular}}
 \label{tab:gridsearchnextrefdataset}
\end{table}

\subsection{Example of predictions}
\label{sec:referencedatasetpredictionexample}

\begin{figure*}
\centering
  \includegraphics[width=0.49\linewidth]{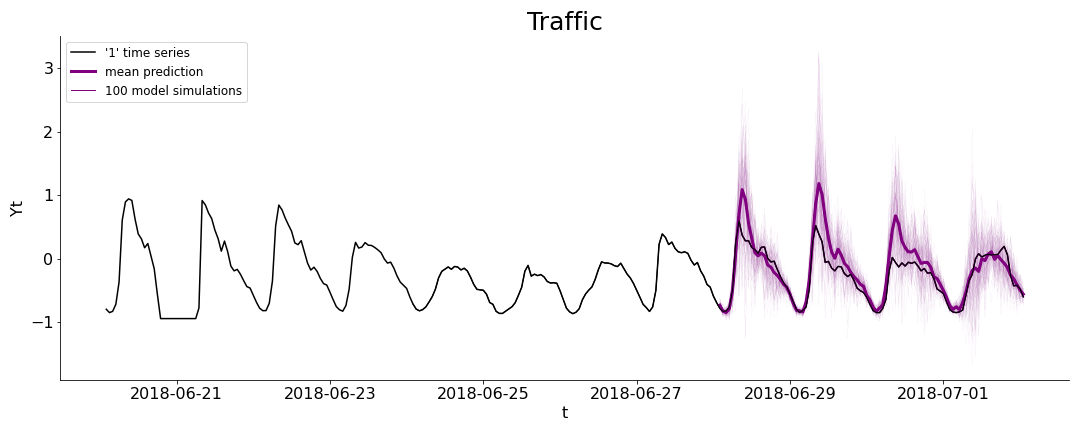}
  \includegraphics[width=0.49\linewidth]{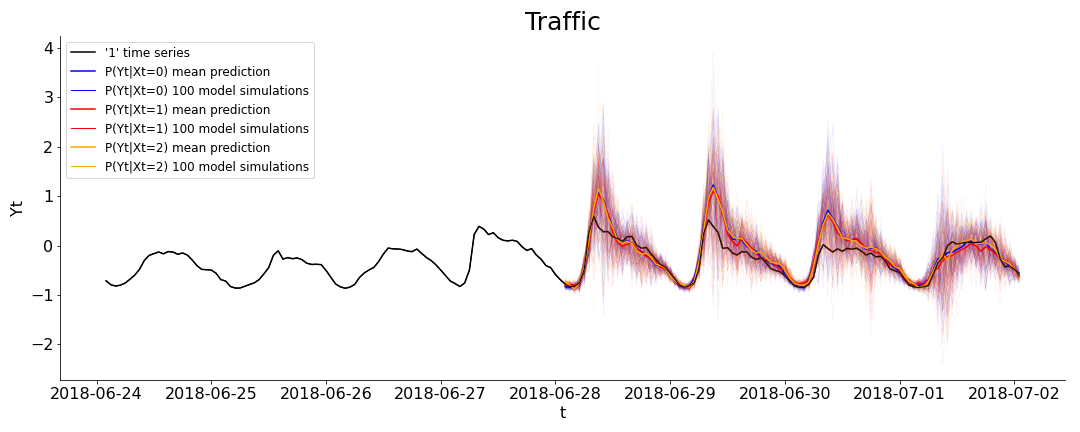}
  \includegraphics[width=0.49\linewidth]{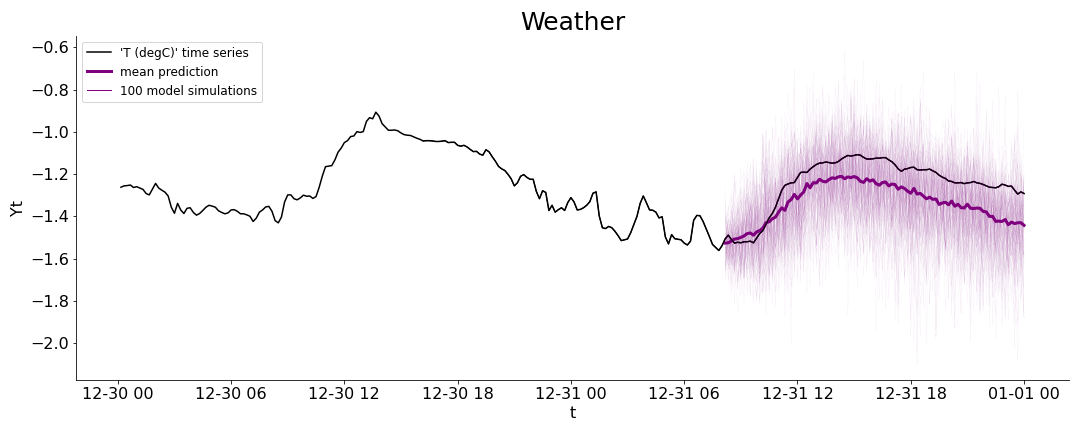}
  \includegraphics[width=0.49\linewidth]{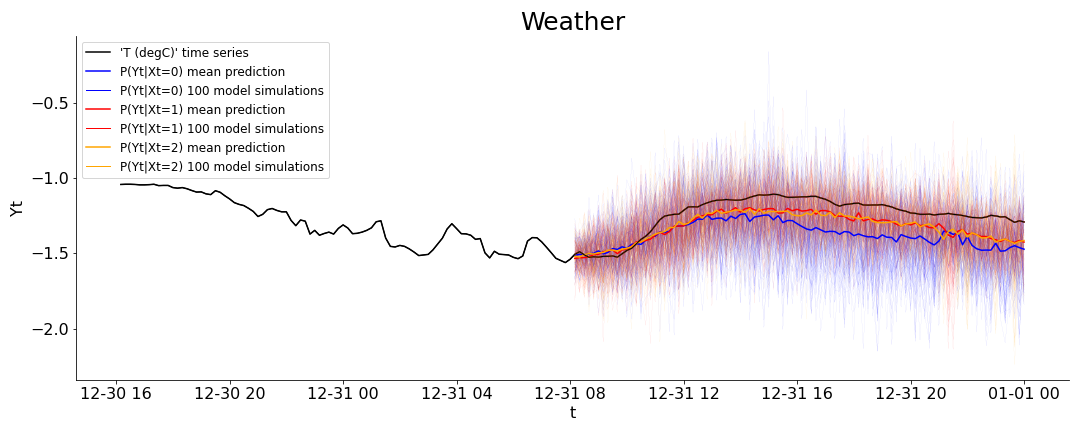}
  \includegraphics[width=0.49\linewidth]{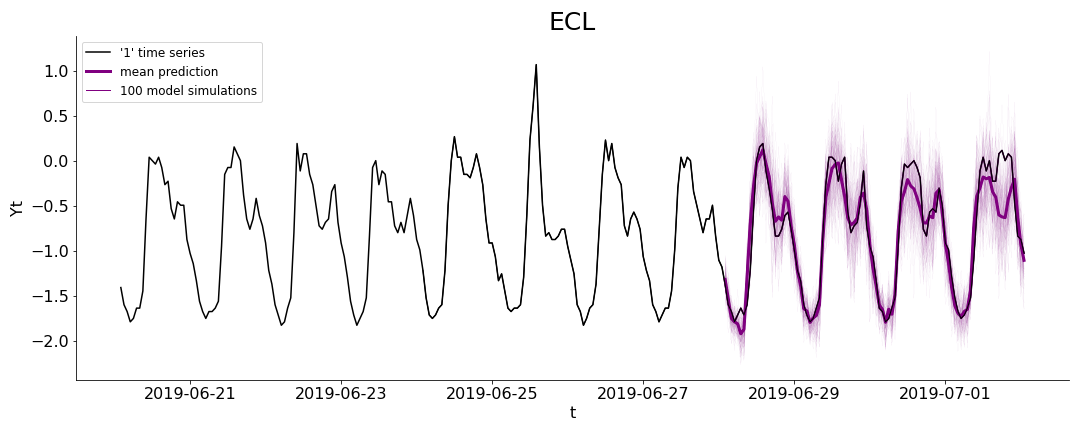}
  \includegraphics[width=0.49\linewidth]{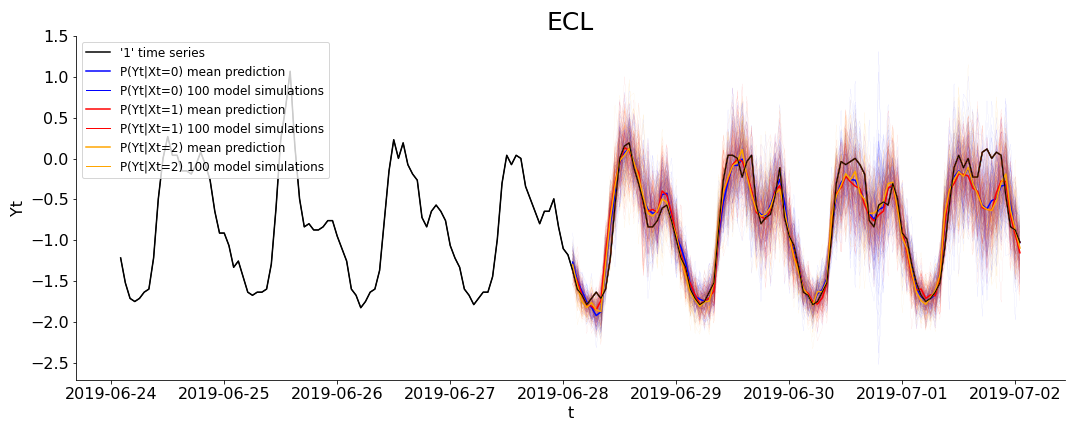}
  \includegraphics[width=0.49\linewidth]{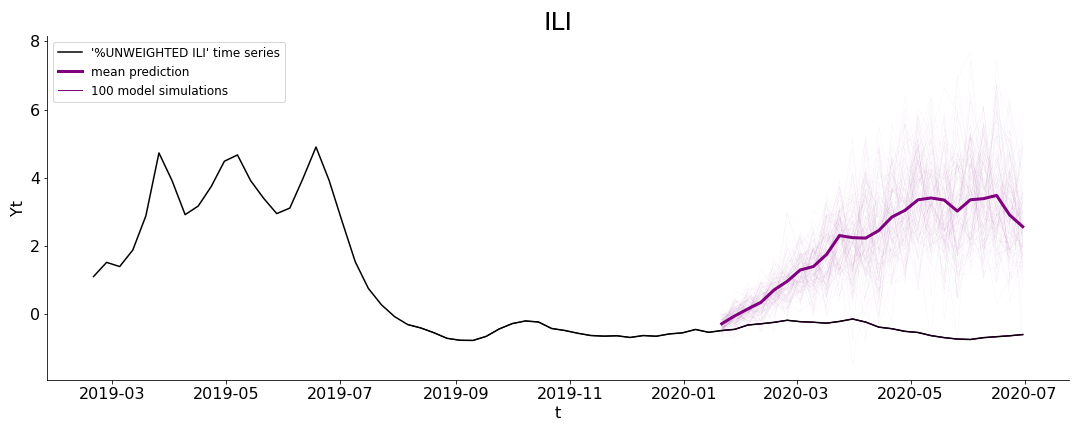}
  \includegraphics[width=0.49\linewidth]{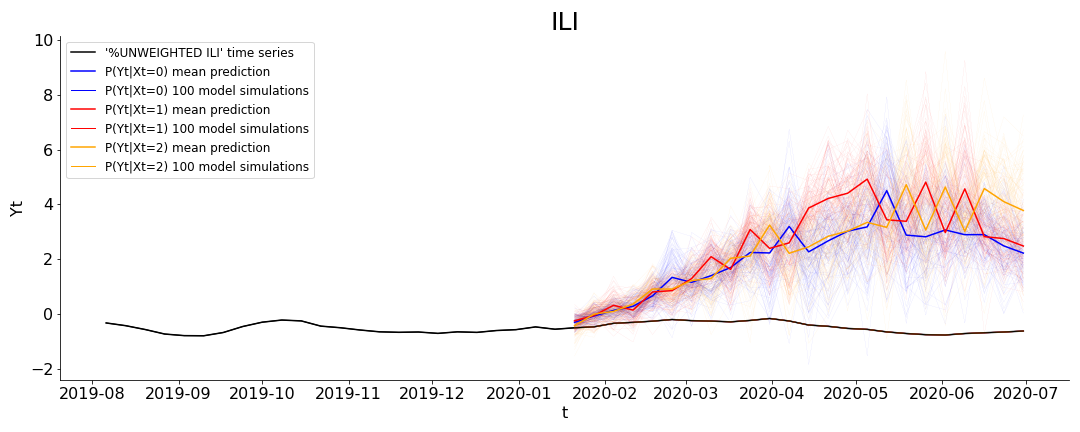}
\caption{\textbf{Predictions on reference datasets.} Example of final prediction of the proposed model on the reference datasets Traffic, Weather, ECL and ILI. (Left) 100 simulations along with the mean prediction of the model (Right) 100 simulations and mean prediction of the three emission laws when the hidden state is fixed to 0, 1 or 2.}
\label{fig:nextpredictionexamplerefdataset1}
\end{figure*}

\begin{figure*}
\centering
  \includegraphics[width=0.49\linewidth]{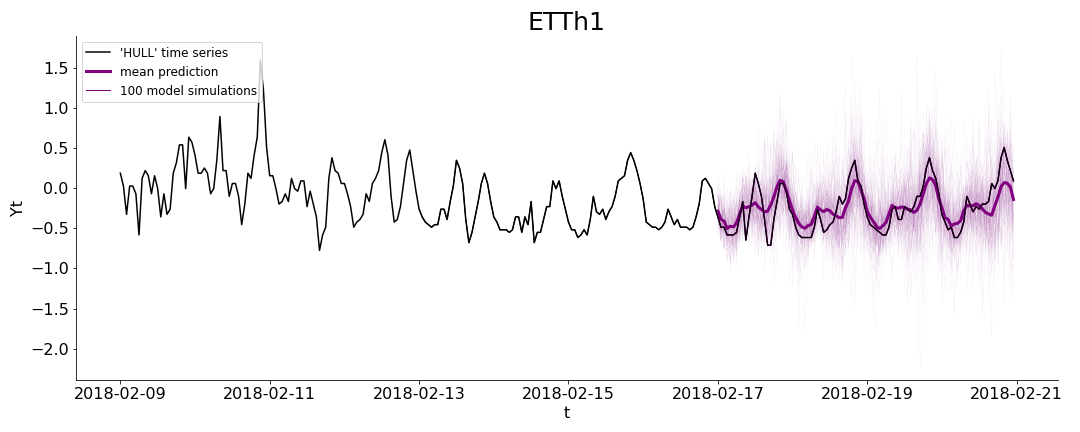}
  \includegraphics[width=0.49\linewidth]{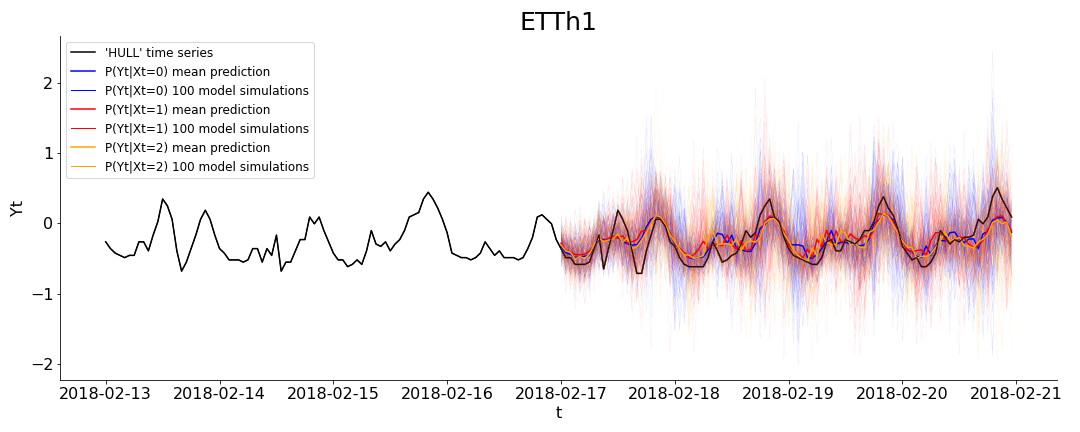}
  \includegraphics[width=0.49\linewidth]{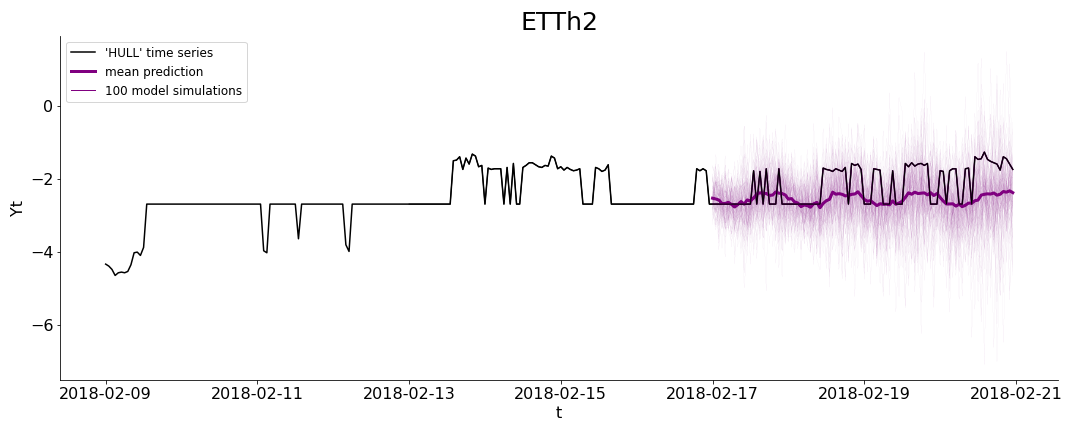}
  \includegraphics[width=0.49\linewidth]{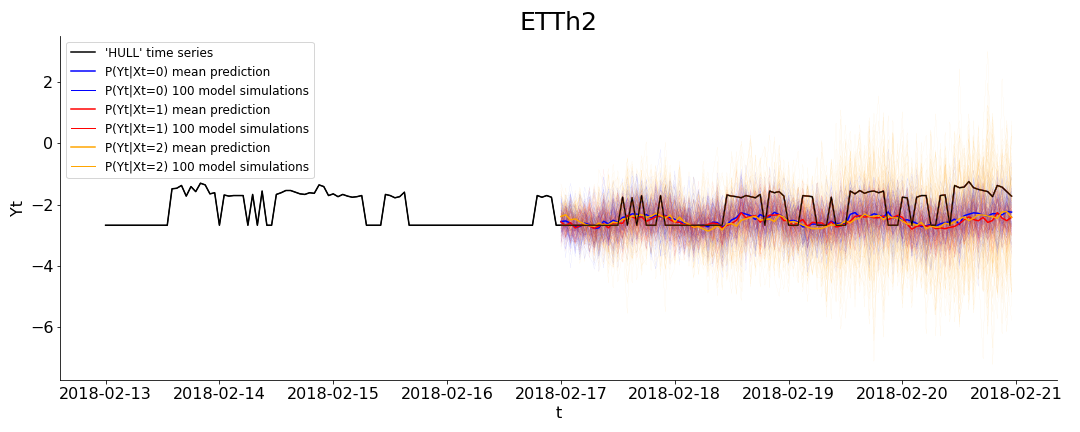}
  \includegraphics[width=0.49\linewidth]{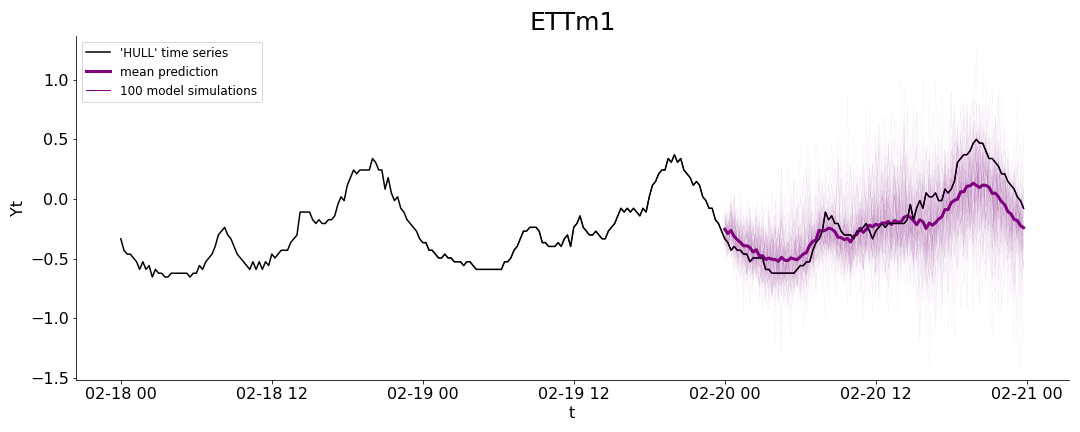}
  \includegraphics[width=0.49\linewidth]{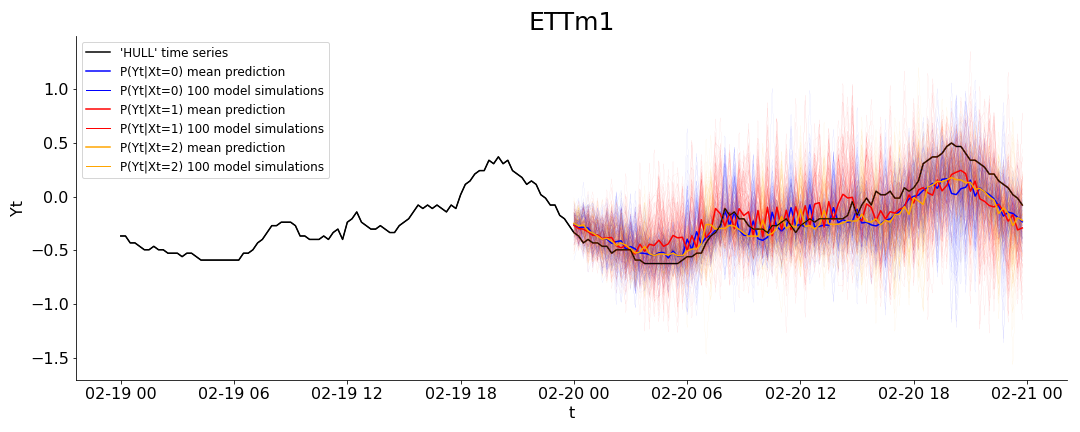}
  \includegraphics[width=0.49\linewidth]{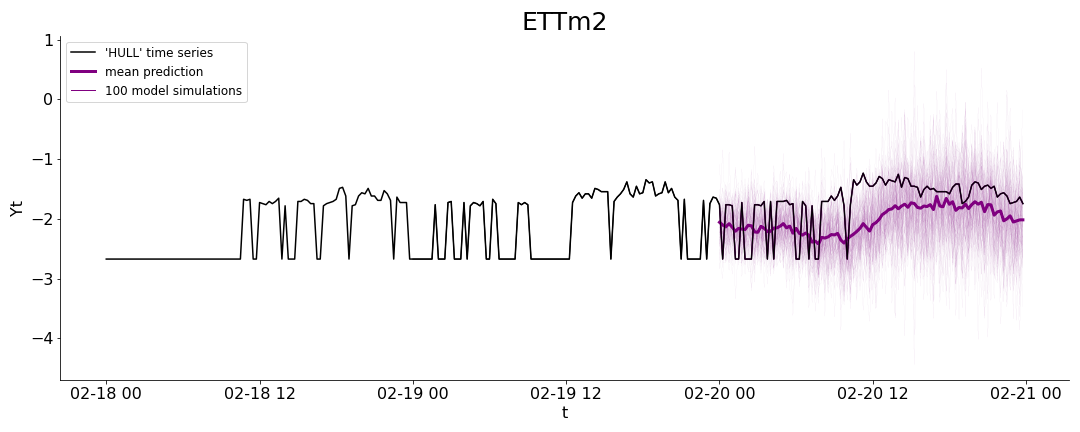}
  \includegraphics[width=0.49\linewidth]{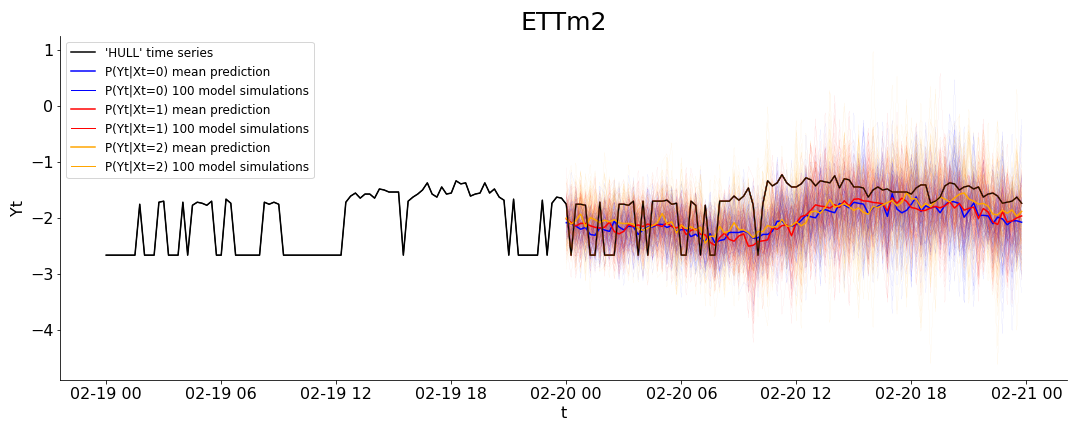}
\caption{\textbf{Predictions on reference datasets.} Example of final prediction of the proposed model on the reference datasets ETTh1, ETTh2, ETTm1 and ETTm2. (Left) 100 simulations along with the mean prediction of the model (Right) 100 simulations and mean prediction of the three  emission laws when the hidden state is fixed to 0, 1 or 2.}
\label{fig:nextpredictionexamplerefdataset2}
\end{figure*}

Finally, we provide examples of predictions on the 8 reference datasets. Figure~\ref{fig:nextpredictionexamplerefdataset1} and \ref{fig:nextpredictionexamplerefdataset2} display for each dataset a prediction of the proposed approach along with the prediction of the emission laws when the hidden state is fixed to 0, 1 or 2.

\end{document}